\documentclass[11pt, a4paper, logo, copyright, nonumbering]{antgroup}
\usepackage[square, numbers]{natbib}
\usepackage{dblfloatfix}
\usepackage{ulem}
\usepackage{caption}
\usepackage{dramatist}
\usepackage{xspace}
\usepackage{pifont}
\usepackage{multirow}
\usepackage{float}
\usepackage{placeins}
\usepackage{makecell}
\usepackage{threeparttable} 

\usepackage{xltabular}
\usepackage{longtable}
\usepackage{hyperref}
\usepackage{amsfonts}
\usepackage{amsmath}
\usepackage{amssymb}
\usepackage{lineno}
\usepackage{multirow}
\usepackage{adjustbox}

\usepackage[bottom]{footmisc}

\usepackage{CJKutf8}
\usepackage{setspace}
\usepackage{dsfont}
\usepackage{array}
\usepackage{tabularx}
\usepackage{xcolor}
\usepackage{tabularx}
\usepackage{booktabs}
\usepackage{subcaption}

\usepackage{algorithm}
\usepackage{algpseudocode}
\usepackage{amsmath}
\usepackage{amsfonts}

\usepackage{booktabs} 
\usepackage{amsmath}  

\usepackage{lipsum}
\usepackage{multicol}
\usepackage{authblk}
\usepackage{wrapfig}
\usepackage[most,skins,theorems]{tcolorbox}
\usepackage{array}
\usepackage{siunitx}
\usepackage{pifont}
\definecolor{customgray}{RGB}{230,230,230}

\usepackage{graphicx}
\usepackage{subcaption}
\usepackage{float}  

\tcbset{
  aibox/.style={
    width=\linewidth,
    top=8pt,
    bottom=4pt,
    colback=gray!10!white,
    colframe=gray!50!black,
    colbacktitle=gray!70!black,
    enhanced,
    center,
    attach boxed title to top left={yshift=-0.1in,xshift=0.15in},
    boxed title style={boxrule=0pt,colframe=white,},
  }
}
\newtcolorbox{AIbox}[2][]{aibox,title=#2,#1}

\makeatletter
\def\@BTrule[#1]{%
  \ifx\longtable\undefined
    \let\@BTswitch\@BTnormal
  \else\ifx\hline\LT@hline
    \nobreak
    \let\@BTswitch\@BLTrule
  \else
     \let\@BTswitch\@BTnormal
  \fi\fi
  \global\@thisrulewidth=#1\relax
  \ifnum\@thisruleclass=\tw@\vskip\@aboverulesep\else
  \ifnum\@lastruleclass=\z@\vskip\@aboverulesep\else
  \ifnum\@lastruleclass=\@ne\vskip\doublerulesep\fi\fi\fi
  \@BTswitch}
\makeatother

\addto\extrasenglish{
}

 {\begin{list}{}%
         {\setlength{\leftmargin}{#1}}%
         \item[]%
 }
 {\end{list}}

\reportnumber{001} %

\renewcommand{\today}{}

\title{\centering LingDT-VL-OCR: Structure-Aware Document-Level Parsing with Fine-Grained Visual Reference}

\author{
   Siyi Qian$^{*}$
   \;\; Xiongfei Bai$^{*}$
   \;\; Bingtao Fu
   \;\; Yichen Lu
   \;\; Gaoyang Zhang
   \;\; Xudong Yang
   \;\; Peng Zhang$^{\dagger}$\\
   Ant Group\\
   {\tt\small \{\href{mailto:siyi.qsy@antgroup.com}{siyi.qsy}, \href{mailto:baixiongfei.bxf@antgroup.com}{baixiongfei.bxf}, \href{mailto:biaotao.bf@antgroup.com}{biaotao.bf}, \href{mailto:luyichen.lyc@antgroup.com}{luyichen.lyc}, \href{mailto:zgy528076@antgroup.com}{zgy528076}, \href{mailto:jiegang.yxd@antgroup.com}{jiegang.yxd}, \href{mailto:minghua.zp@antgroup.com}{minghua.zp}\}@antgroup.com}
}

\footnotetext[1]{Equal contribution.}
\footnotetext[2]{Corresponding to minghua.zp@antgroup.com}

\renewcommand{\phi}{\varphi}

\renewcommand{\epsilon}{\varepsilon}
\renewcommand{\imath}{\mathrm{i}}

\newlength{\restsubwidth}
\newlength{\restsubheight}
\newlength{\restsubmoreheight}
\newcommand{\rest}[2]{%
        \settowidth{\restsubwidth}{\ensuremath{#2}}
        \settoheight{\restsubheight}{\ensuremath{{}_{#2}}}
        \ensuremath{{#1\hskip 0.5pt}_{\vrule\kern2pt\parbox[b][%
        4pt][b]{\the\restsubwidth}{%
                        \ensuremath{{}_{#2}}}}}
        }

\begin{abstract}

In this paper, we propose \textbf{LingDT-VL-OCR}, a document parsing system tailored to financial-domain documents, transforming ultra-long financial PDFs into semantically consistent, highly accurate, structured outputs with auditing-grade provenance. To address finance-specific challenges such as complex layouts, cross-page structural discontinuities, and cell-level referencing capability, \textbf{LingDT-VL-OCR} combines (1) a Cross-page Contents Consolidation algorithm to restore continuity across pages and a Document-level Heading Hierarchy Reconstruction (DHR) module to build a globally consistent Table of Contents (TOC) tree for structure-aware retrieval, and (2) a difficulty-adaptive curriculum learning training strategy for table parsing, together with a CellBBoxRegressor module that uses structural anchor tokens to localize table cells from decoder hidden states without external detectors. Experiments demonstrate that our model shows high performance on the Overall metric of OmniDocBench. To enable realistic evaluation in the financial vertical, we further introduce \textbf{FinDocBench}, a benchmark that includes six financial document categories with expert-verified annotations and evaluation metrics including Table of Contents edit-distance-based similarity (TocEDS), cross-page concatenated TEDS, and Table Cell Intersection over Union (C-IoU). We evaluate a wide range of state-of-the-art models on \textbf{FinDocBench} to assess their capabilities and remaining limitations on financial documents. Overall, \textbf{LingDT-VL-OCR} and \textbf{FinDocBench} provide a practical foundation for reliable downstream financial document applications.

\end{abstract}

\begin{document}
\maketitle

\newpage

\section{Introduction}

Document parsing~\cite{zhang2024document, li2025monkeyocr, niu2025mineru2, feng2025dolphin} extracts machine-readable data from unstructured files like PDFs. As a core component of document intelligence, it is essential for powering Retrieval-Augmented Generation (RAG)~\cite{lewis2020retrieval, gao2023retrieval, jiang2023active} pipelines with structured, reliable context. Although recent research has explored a variety of approaches, including pipeline-based frameworks, multi-stage systems, and Multimodal Large Language Model (MLLM)~\cite{achiam2023gpt, qwen2.5, bai2025qwen3} end-to-end solutions. Although existing frameworks have already achieved substantial results in page-level analysis, most document parsing models and benchmarks~\cite{zhong2019publaynet, pfitzmann2022doclaynet, ouyang2025omnidocbench} continue to focus predominantly on treating pages as isolated entities. This narrow focus fails to capture the global logical flow and hierarchical structure of long documents, leading to semantic fractures and a loss of vital context.

In the financial and insurance domain, document parsing must satisfy much stricter business requirements, including accurate understanding of business scenarios, high-precision information extraction, and audit traceability. Unlike general documents, financial materials are characterized by their vast diversity and extreme length, ranging from A-share, Hong Kong, and US annual reports to audit reports and prospectuses that often span hundreds of pages. Despite the progress in existing methods, they have seen limited practical adoption in real financial scenarios due to their insufficient alignment with industry requirements. Financial document parsing introduces several unique challenges:

\begin{itemize}
    \item \textbf{Layout Complexity}: Financial documents often use multi-column designs; standard parsers frequently read across columns, incorrectly merging unrelated text and destroying semantic meaning.
    \item \textbf{Semantic Fragmentation and Hierarchical Disconnect:} Financial documents are intrinsically governed by a rigorous, multi-level structural logic. However, physical pagination forcefully severs headers from their corresponding clauses, fragmenting this global hierarchy. When downstream applications like RAG and DocQA~\cite{mathew2021docvqa, mathew2022infographicvqa} rely on isolated page-level parsing, they suffer from severe context loss.
    \item \textbf{Visual Reference Requirements}: Unlike general documents, financial institutions require reliable source referencing over intricate tables like financial statements, requiring every data point to be mapped back to precise page coordinates or specific table cells for visual auditing and compliance.
    \item \textbf{Absence of Financial Document Benchmark}: Existing document parsing benchmarks focus on general or academic contents, failing to represent the unique structural challenges and precision needs of the financial sector.
\end{itemize}

To comprehensively evaluate these capabilities under financial scenarios, we further construct \textbf{FinDocBench}, a document parsing benchmark oriented to financial vertical. FinDocBench comprises six distinct sub-categories of financial documents, accompanied by meticulously human-labeled and expert-verified annotations. As mentioned above, the data embodies the characteristics of financial documentation: extreme length, hierarchical headings, complex layouts, intricate cross-page tables and precise table cell localization. Alongside the dataset, we further introduce a corresponding evaluation pipeline to measure model capability when facing the mentioned challenges, including a new metric, \textbf{Table of Contents edit-distance-based similarity} (TocEDS), tree edit-distance-based similarity (TEDS) of concatenated cross-page tables and normalized edit distance (NED).


In summary, this work presents a financial scenario oriented document parsing system designed to meet the strict, precise, and auditable requirements of real‑world financial enterprise workflows.
Our contributions are as follows:

\begin{itemize}
    \item \textbf{Document-level Parsing.} To handle the structural complexity of financial documents, we propose a document-level content reconstruction algorithm with two core components: cross-page contents consolidation, which semantically concatenates cross-page text blocks and tables; and document-level heading hierarchy reconstruction, which reassigns heading levels across the entire document to ensure structural consistency.

    \item \textbf{Precision Table Parsing with Visual Reference.} To address the structural complexity of financial tables, we adopt a curriculum based difficulty adaptive sampling to ensure table parsing accuracy; and a cell-level visual reference that localizes cells via structural anchor token hidden states and regresses them into bounding boxes to enable finance-grade visual auditing for compliance-critical workflows.

    \item \textbf{Financial vertical document parsing benchmark: FinDocBench.} To evaluate parsing models in real-world financial scenarios, we introduce FinDocBench, covering diverse documents with quintessential financial characteristics and the corresponding evaluation pipeline. 
\end{itemize}

\section{Related Work}

This section surveys the existing landscape of document parsing technologies, covering general-purpose OCR frameworks~\cite{cui2025paddleocr3, cui2025paddleocr, niu2025mineru2, team2025hunyuanocr}, layout analysis models~\cite{zhao2024doclayoutyoloenhancingdocumentlayout}, table structure recognition methods~\cite{Zhong2019ImagebasedTR, smock2022pubtables}, and multimodal large language models for document understanding~\cite{wang2024docllm, ye2023mplug}. We identify key limitations of current approaches when applied to financial scenarios: single-page processing, absence of cross-page semantic linking, and no audit traceability.

\subsection{Traditional OCR Pipelines}
Early approaches to document parsing primarily relied on modular pipelines, which decouple the process into text detection, recognition, and layout analysis. Traditional engines like Tesseract~\cite{smith2007overview} and ABBYY FineReader~\cite{abbyy_finereader_software} laid the foundation by focusing on character-level accuracy. More recently, frameworks such as PaddleOCR~\cite{cui2025paddleocr3} have gained significant traction. Specifically, PaddleOCR’s PP-OCR series~\cite{du2020pp} provides a robust, lightweight pipeline for multi-language recognition, while its PP-Structure module~\cite{li2022pp} introduces sophisticated tools for layout analysis and table recovery.

Furthermore, the integration of deep learning led to the development of layout-aware models like LayoutLM (v1-v3)~\cite{xu2020layoutlm, xu2021layoutlmv2, huang2022layoutlmv3} and LayoutXLM~\cite{xu2021layoutxlm}, which utilize spatial coordinates and visual features as additional embeddings. While these models excel at identifying localized components (e.g., headers, footers, or individual table cells), they still operate within a ``detect-then-recognize'' paradigm. This modularity often leads to error accumulation across stages—where a minor detection flaw propagates into a semantic error. Crucially, even advanced frameworks like PaddleOCR are typically optimized for single-page processing. In financial contexts, they often fail to maintain semantic coherence when a single table spans multiple pages or when a financial term requires a cross-page definition, and they lack the intrinsic referencing capability needed to link extracted data back to its specific pixel-level origin for auditing purposes.

\subsection{General Vision Language Models}
The emergence of Vision Language Models (VLMs), such as commercial GPT series~\cite{achiam2023gpt}, Gemini family~\cite{team2023gemini} and open-source Qwen collection~\cite{wang2024qwen2,qwen2.5, bai2025qwen3}, has demonstrated remarkable zero-shot capabilities in document reasoning. These models treat document images as visual inputs and leverage the linguistic power of LLMs to interpret contents. However, general VLMs are often criticized for their ``hallucination'' tendencies regarding numerical data~\cite{chen2021finqa, li2023evaluating}, which is a critical failure point for financial auditing. Moreover, their architectural design is usually optimized for natural scenes rather than document-specific features like small-font footnotes or complex cross-cell dependencies in tables.

\subsection{OCR Specialized Vision Language Models}
To bridge the gap between general vision and precise text extraction, recent research has pivoted towards OCR-specialized VLMs. Models like Donut~\cite{kim2022ocr} and Nougat~\cite{blecher2023nougat} proposed an end-to-end Transformer architecture that maps document images directly to structured markups (e.g., JSON or LaTeX) without intermediate OCR steps. More recently, models such as Vary~\cite{wei2024vary} and GOT-OCR~\cite{wei2024general} have enhanced the resolution and granularity of visual encoders to handle high-density text. Despite their progress, these models are predominantly designed for single-page or snippet-level processing. They lack the mechanism to handle long-form financial documents where semantic coherence must be maintained across dozens of pages.


\section{Method}

\subsection{Overall Framework}

\begin{figure}[t]
    \centering
    \includegraphics[width=\linewidth]{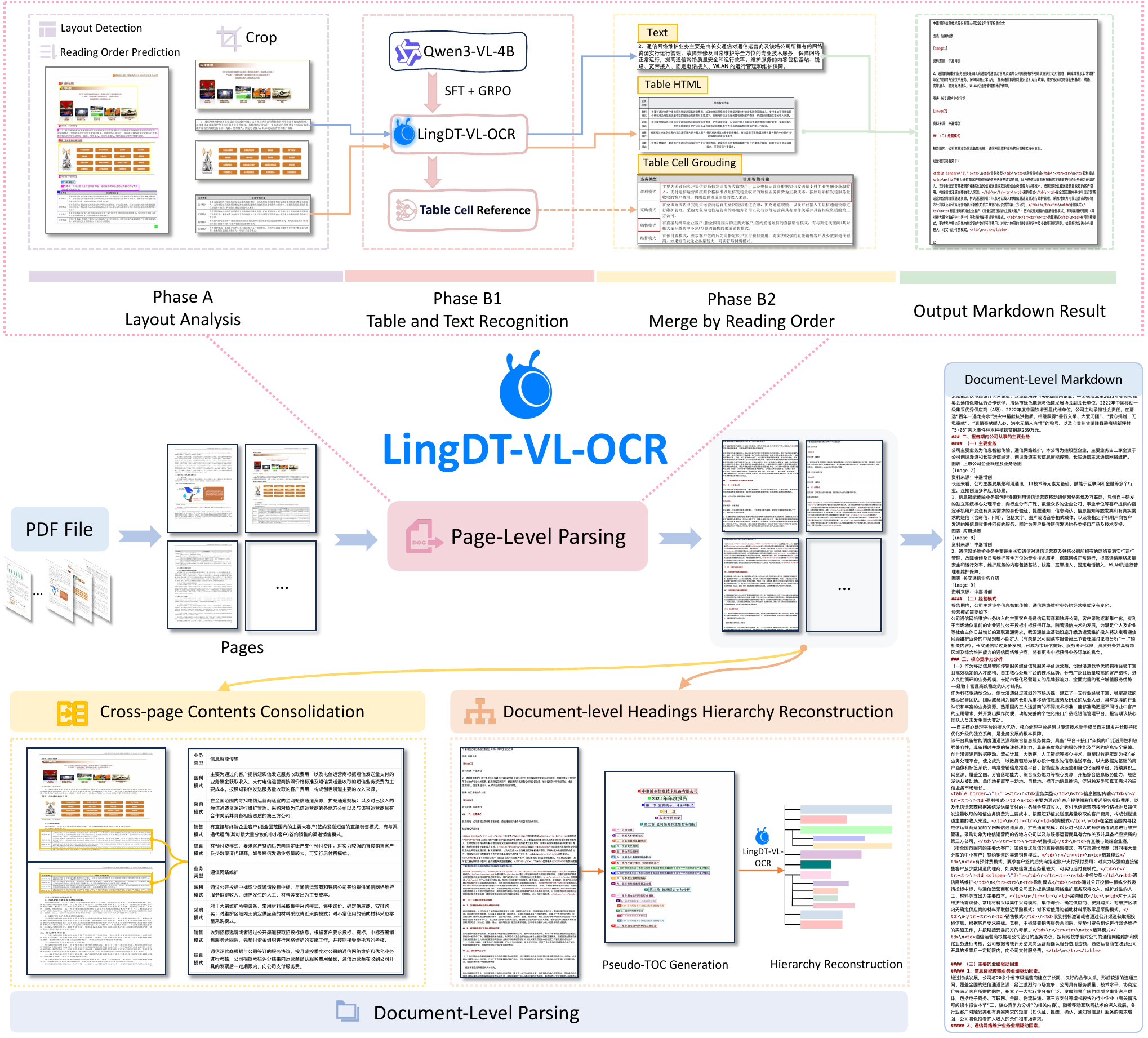}
    \caption{\textbf{LingDT-VL-OCR overall architecture.}}
    \label{fig:pipeline}
\end{figure}

As show in Figure~\ref{fig:pipeline}, our framework transforms complex, unstructured financial documents into high-fidelity, machine-readable formats while maintaining a rigorous, end-to-end audit trail. To preserve semantic and structural continuity across fragmented layouts, the pipeline integrates cross-page consolidation with a Document-level Heading Hierarchy Reconstruction (DHR) module, which leverages multimodal features to reconstruct a coherent document-level heading hierarchy, serving as the document's unified structural skeleton. For dense data extraction, we employ curriculum-based table parsing to adaptively handle structural complexity and utilize a novel ``structural anchor token'' mechanism to enable direct cell-level spatial grounding from decoder hidden states. By mapping extracted data units back to their original visual coordinates without auxiliary detectors, this unified approach ensures the precision, narrative integrity, and empirical referencing capability required for professional financial table auditing.

\subsection{Cross-page Contents Consolidation}
To ensure linguistic and structural continuity, we implement a consolidation mechanism that transforms disjointed pages into a fluid semantic content. This process employs semantic restoration logics for unstructured text and structured tables:
\begin{itemize}
    \item \textbf{Cross-page Text Merge}: The system identifies cross-page text fragments at page boundaries, stripping away non-content elements such as headers and footers before merging the body text with the leading contents of the subsequent page to maintain syntactic and contextual integrity.
    \item \textbf{Cross-page Table Merge}: 
To ensure the structural integrity of long financial document with cross-page tables, we implement an adaptive heuristic-based splicing mechanism. A table fragment $T_n$ on page $n$ is evaluated for merging with $T_{n+1}$ on the succeeding page based on three hierarchical criteria:

\begin{itemize}
    \item[(i)] \textbf{Structural Alignment}: The column dimensions must be strictly consistent, i.e., $\text{Dim}(T_n.cols) = \text{Dim}(T_{n+1}.cols)$.
    \item[(ii)] \textbf{Contextual Proximity}: To ensure physical continuity, the set of intervening semantic elements $E$ between fragments (excluding non-content artifacts such as page headers and footers) must be empty($E = \emptyset$).This condition confirms that the table fragments are not separated by unrelated discourse or structural interruptions.
    \item[(iii)] \textbf{Adaptive Header Splicing}: Upon satisfying (i) and (ii), the merging strategy adapts to the header content:
    \begin{itemize}
        \item[\textit{(a)}] \textbf{Homogeneous Splicing}: If $T_{n+1}$ lacks a header or possesses one identical to $T_n$, it is treated as a seamless continuation. The redundant header is discarded, and only the \texttt{<tbody>} is appended.
        \item[\textit{(b)}] \textbf{Heterogeneous Splicing}: If $T_{n+1}$ contains a distinct header, it indicates a categorical transition within the same structural entity. The entire fragment $T_{n+1}$ is merged into $T_n$ to preserve the sub-header information.The table cases are shown in Appendix~\ref{app:Cross-page table cases}.
        
    \end{itemize}
\end{itemize}

The formal reconstruction procedure is detailed in Algorithm \ref{alg:merge_logic}.

\begin{algorithm}[H]
\caption{Adaptive Cross-page Table Reconstruction}
\label{alg:merge_logic}
\begin{algorithmic}[1]
    \Require Sequence of table fragments $\mathcal{T} = \{T_1, T_2, \dots, T_k\}$
    \Ensure A set of unified tables $\mathcal{M}$
    
    \State $\mathcal{M} \gets \emptyset, T_{anchor} \gets T_1$
    
    \For{$i = 2$ \textbf{to} $k$}
        \State $T_{next} \gets T_i$
        \State \textit{// (i) \& (ii) Preliminary Constraint Check}
        \State $C_{align} \gets (\text{Dim}(T_{anchor}.cols) = \text{Dim}(T_{next}.cols))$
        \State $E \gets \text{GetInterveningElements}(T_{anchor}, T_{next})$
        
        \If{$C_{align} \land (E = \emptyset)$}
            \State \textit{// (iii) Adaptive Header Logic}
            \If{$\neg \text{HasHeader}(T_{next}) \lor (\text{TableHead}(T_{next}) = \text{TableHead}(T_{anchor}))$}
                \State \textit{// Case a: Seamless Merge (Append rows only)}
                \State $T_{anchor} \gets \text{MergeBody}(T_{anchor}, T_{next})$
            \Else
                \State \textit{// Case b: Direct Full Merge (Append entire table structure)}
                \State $T_{anchor} \gets \text{MergeFullTable}(T_{anchor}, T_{next})$
            \EndIf
        \Else
            \State $\mathcal{M} \gets \mathcal{M} \cup \{T_{anchor}\}$
            \State $T_{anchor} \gets T_{next}$
        \EndIf
    \EndFor
    \State $\mathcal{M} \gets \mathcal{M} \cup \{T_{anchor}\}$
    \State \Return $\mathcal{M}$
\end{algorithmic}
\end{algorithm}

\end{itemize}



\subsection{Document-Level Heading Hierarchy Reconstruction}
Real-world financial documents often exhibit complex heading
hierarchies that span dozens to hundreds of pages.
This
significantly undermines downstream tasks performance due to severe semantic
fragmentation caused by physical pagination.
To solve this problem,
We propose a module for document hierarchy structure extraction,
dubbed Document-level Heading Hierarchy Reconstruction (DHR).
DHR transforms a sequence of isolated page-level headings into a globally
consistent Table of Contents (TOC) tree, which serves as a structural
abstraction for the entire document, fueling downstream tasks like RAG and
document QA. The MMRAG-DocQA framework~\cite{gong2025mmrag} validates that hierarchical indices integrating cross-page dependencies are essential for high-quality retrieval in multi-page documents
. Similarly, HiChunk~\cite{lu2025hichunk} proves that multi-level document structuring and auto-merge retrieval algorithms significantly enhance the overall reasoning precision and quality of downstream RAG systems.
As we will show in Section~\ref{sec:result-dhr}, DHR yields substantial
improvements on longer documents, while providing comparable performance on
shorter documents where simpler structures are already well-captured by text
alone.

In this section,
we first formalize the notations and define the goal of the DHR module.
We then propose its core component that achieves this TOC tree abstraction.
Finally, we demonstrate its impact on document intelligence applications.

\subsubsection{Notations}
Given a document $\mathcal{D}$, we denote $\mathcal{H} = \{h_1, h_2, \dots,
h_n\}$ as the ordered sequence of candidate headings within $\mathcal{D}$.
We can then characterize an entry $h_i$ in this sequence using 4 attributes:
\begin{equation}
    h_i = \langle
        \mathcal{T}_i, \mathcal{V}_i, \mathcal{S}_i, \ell_i
    \rangle, \; \text{where} \label{eq:dhr-formalization}
\end{equation}
\begin{itemize}
    \item
        $\mathcal{T}_i$ is the raw textual content of the heading.
    \item
        $\mathcal{V}_i$ denotes the visual appearance of the header, reflecting latent typographic properties like font weight, style, and point size.
    \item
        $\mathcal{S}_i$ is the spatial coordinates (bounding box) and page-index
        anchor.
    \item
        $\ell_i \in \{1, \dots, L\}$ is the heading's level (e.g., $\ell=1$
        for ``Chapter'', $\ell=2$ for ``Section'', etc.).
\end{itemize}
Among the 4 attributes listed above, $\ell_i$ encodes the heading hierarchy.
However, only $\mathcal{T}_i$, $\mathcal{V}_i$, and $\mathcal{S}_i$ are
available from upstream layout analysis results.
We therefore propose the DHR module to recover $\ell_i$,
effectively reconstructing the heading hierarchy of the whole document.

\subsubsection{Reconstruction Pipeline}

\begin{figure}
    \centering
    {
        \newcommand{\debugwidth}[1]{#1}
        \renewcommand{\debugwidth}[1]{{\setlength{\fboxsep}{0pt}\fbox{#1}}}
        \renewcommand{\debugwidth}[1]{#1}
        \debugwidth{\begin{minipage}[t]{.66\linewidth}
            \centering
            \subcaption*{\quad\, Pseudo-TOC Aggregation}
        \end{minipage}}%
        \debugwidth{\begin{minipage}[t]{.33\linewidth}
            \centering
            \subcaption*{\hspace{.7em} VLM Prompting}
        \end{minipage}}
        \includegraphics[width=0.93\linewidth]{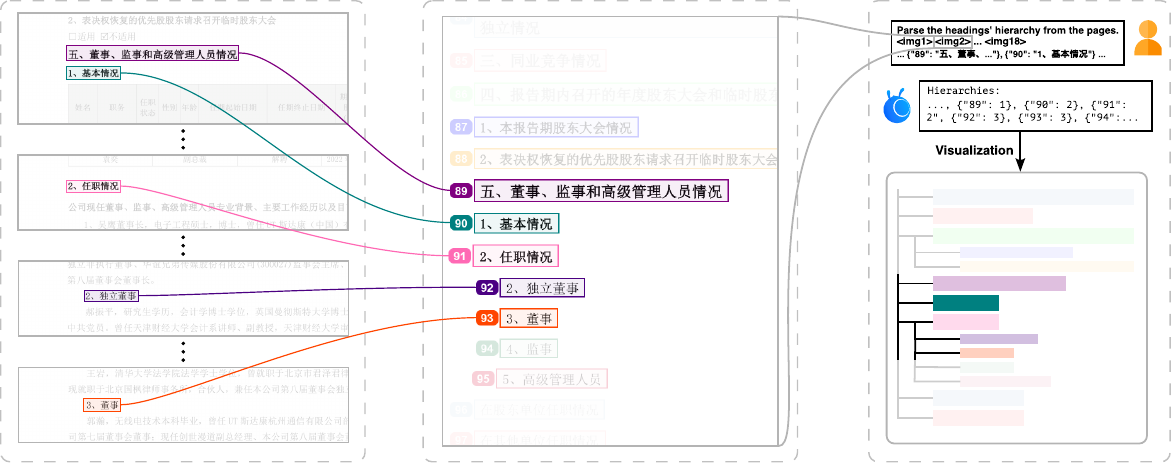}
    }
    \caption{
        \textbf{Conceptual overview of the DHR pipeline.}
        Headings are cropped from the original documents (\textbf{left}) and
        aggregated into a pseudo-TOC (\textbf{middle}).
        The pseudo-TOC is then fed to a VLM, along with available textual
        information (Eq. \ref{eq:dhr-formalization}), obtaining the heading
        hierarchy (\textbf{right}).
    } \label{fig:toc-tree}
\end{figure}

The proposed DHR module recovers the heading hierarchy of the entire document
with the help of the multi-modal reasoning ability of Fin-OCR.
As illustrated in Figure~\ref{fig:toc-tree}, the module first extracts headings
from the original document to form a set of images that resemble a pseudo-TOC,
then it feeds the available textual results and the pseudo-TOC into Fin-OCR to
reconstruct the document's heading hierarchy.
We detail the process in the following.

\paragraph{Pseudo-TOC Aggregation.}
We find the layout analysis results on the original documents to provide a solid
ground for accurate hierarchy reconstruction.
Specifically, this includes:
\begin{itemize}
    \item \textbf{Textual semantics from $\mathcal{T}_i$:}
        A heading $h_i$'s text content contextualizes itself in the whole
        document.
        When provided as additional text prompts to a VLM, it steers the VLM for
        more robust headings understanding (e.g., ``Note 5.3'' often follows
        ``Note 5.2''), resolving potential ambiguities.
    \item \textbf{Visual style from $\mathcal{V}_i$:}
        In financial reports, the heading level is often indicated by consistent
        typography (e.g., all top-level headings share the same font size and
        weight, and larger-scope headings are often formatted with more
        prominent typesetting).
        In a long sequence of headings, $\mathcal{V}_i$ provides helpful cues to
        mitigate level drift.
    \item \textbf{Spatial information from $\mathcal{S}_i$:}
        Bounding box coordinates from the layout analysis encode the
        indentation pattern of headings,
        which provide additional cues for deeply nested heading hierarchies.
\end{itemize}

To effectively preserve these cues,
we directly crop the document pages using each heading's grounded bounding box
coordinates available from the previous layout analysis stage, obtaining an
image crop $\mathcal{C}_i$ for each heading $h_i$ (Figure~\ref{fig:toc-tree}
left).
These image crops are then pasted onto an empty page-sized image, preserving
their horizontal offset from the original document.
By laying out all headings vertically, we can obtain a set of page-sized images,
forming a \textit{pseudo-TOC} (Figure~\ref{fig:toc-tree} middle). 
Inspired by widely-used practices from the GUI agents
literature~\cite{browser_use2024},
we also annotate the pasted heading crops with colored bounding boxes and
corresponding monotonically increasing numeric labels to further boost VLM's
grounding.
This pseudo-TOC is later fed to the VLM and serves as the main source of
heading hierarchy reconstruction.

\paragraph{Document-Level Heading Hierarchy Reconstruction.}
To finally recover the level $\ell_i$ of every heading from the document, we
prompt the
Fin-OCR model with the generated pseudo-TOC images and relevant layout analysis
results (Figure~\ref{fig:toc-tree} top-right).
Specifically, we craft the text-image prompt from the following components:
\begin{itemize}
    \item
        A brief, task-oriented textual instruction. E.g., ``\texttt{Parse the
        TOC pages into the following JSONL format:...}''.
    \item
        The text content $\mathcal{T}_i$ of each heading extracted from the
        layout analysis result and the numeric label of the corresponding
        heading.
    \item
        The pseudo-TOC pages in order as image prompts.
\end{itemize}
The level of each heading is then recovered in the specified format,
the heading hierarchy of the whole document can be easily decoded and visualized
from this format (Figure~\ref{fig:toc-tree} bottom-right).

\subsection{Curriculum Learning and Reinforcement Optimization for Table Parsing}

Difficulty assessment is useful not only for filtering high-quality training data but also for guiding RL optimization. Motivated by this, we collect training data from a mixture of public and in-house sources, and construct a multi-dimensional tabular annotation schema that encompasses structural attributes and stylistic formatting attributes (e.g., table line styles, blank-cell ratio, the maximum rowspan/colspan, and the distribution of rowspan/colspan values). We then build a finely annotated in-house dataset to guide targeted optimization. To identify which attributes materially affect parsing performance, we conduct a Pearson correlation analysis between each attribute \(a\in\mathcal{A}\) and the model’s TEDS score \(y\) on the in-house dataset:
\begin{equation}
\rho(a,y)=\frac{\operatorname{cov}(a,y)}{\sigma_a\,\sigma_y}.
\end{equation}
We then rank attributes by $|\rho(a,y)|$ and treat those with the largest magnitude correlations as the primary difficulty factors. We find that the following two categories of attributes are significantly negatively correlated with the final TEDS: structural complexity (e.g., the values of rowspan/colspan), and the Inference Consistency Difficulty (ICD), defined as the standard deviation of TEDS across repeated high-temperature inference runs for a given sample~\cite{niu2025mineru2}, which serves to quantify instability in structural parsing.

Based on these findings, we define a unified difficulty score \(d(x)\) by combining structural complexity and ICD, and use it to organize training samples for curriculum learning:
\begin{equation}
d(x)=\alpha\,\text{Structural Complexity}(x) + \beta\,\text{ICD}(x).
\end{equation}
We first perform supervised fine-tuning (SFT) on the curriculum-organized samples to establish a strong initialization. We then apply reinforcement learning with Group Relative Policy Optimization (GRPO)~\cite{shao2024deepseekmath} after SFT to improve table row/column alignment. In each iteration, GRPO samples a group of responses $(o_1,\ldots,o_G)$ for a query $q$ from the old policy $\pi_{\theta^{\mathrm{old}}}$, and updates $\pi_\theta$ by maximizing:
\begin{equation}
\begin{aligned}
\mathcal{L}_{\mathrm{GRPO}}(\theta)
&=
\mathbb{E}_{q\sim D,\ \{o_i\}_{i=1}^{G}\sim \pi_{\theta^{\mathrm{old}}}(\,\cdot \mid q\,)}
\Bigg[
\frac{1}{G}\sum_{i=1}^{G}
\Big(
\min\!\big(r_i A_i,\ \mathrm{clip}(r_i,1-\epsilon,1+\epsilon)\,A_i\big)
\\
&\hspace{3.2em}
-\beta\,D_{\mathrm{KL}}\!\big(\pi_\theta(\,\cdot \mid q\,)\,\|\,\pi_{\mathrm{ref}}(\,\cdot \mid q\,)\big)
\Big)
\Bigg],
\qquad
r_i=\frac{\pi_\theta(o_i\mid q)}{\pi_{\theta^{\mathrm{old}}}(o_i\mid q)}.
\end{aligned}
\end{equation}

In our preliminary analysis, we observe that for structurally complex tables, the model may fail to generate row/column-aligned parses, with errors concentrated at the boundaries, in particular the last few rows and the last few columns. Beyond the TEDS reward~\cite{zhang2025trivia}, we explicitly introduce a grid-consistency signal that rewards outputs whose extracted grid signature matches the ground truth, i.e., $\mathbb{I}\!\left[g(o)=g(o^\ast)\right]$.
The reward for each output $o$ is a weighted sum of grid-consistency and TEDS:
\begin{equation}
R(q,o)=
\lambda_1\,\mathbb{I}\!\left[g(o)=g(o^\ast)\right]
+\lambda_2\,\mathrm{TEDS}\!\left(\mathrm{norm}(o),\mathrm{norm}(o^\ast)\right),
\end{equation}
where $g(\cdot)$ extracts the logical grid signature (expanded rowspan/colspan row-width list), and $o^\ast$ is the ground-truth table. The advantage $A_i$ is computed from group rewards using a mean-baseline within the sampled group.

To stabilize training, outputs that exceed the maximum length or violate the required table schema are assigned zero reward.

We apply an ensemble of table parsers (PaddleOCR-VL~\cite{cui2025paddleocr}, Mineru2.5~\cite{niu2025mineru2}, and HunyuanOCR~\cite{team2025hunyuanocr}) for automatic pre-filtering to bootstrap an easy-stage subset with reliable supervision. Building on the difficulty score \(d(x)\), we further stratify the remaining data into different difficulty levels and prioritize samples with higher difficulty for later-stage training. We continuously monitor and adjust the difficulty distribution in each sampling round to construct the advanced-stage subset, and conduct an additional training cycle on these challenging samples, thereby improving table parsing accuracy. With GRPO, the model generates better row/column-aligned parses on complex tables, especially improving alignment in the last few rows and columns. More results are given in Appendix~\ref{app:grpo_alignment}.

\subsection{ Table Parsing with Cell-Level Visual Reference  }

\begin{figure}[t]
    \centering
    \includegraphics[width=\linewidth]{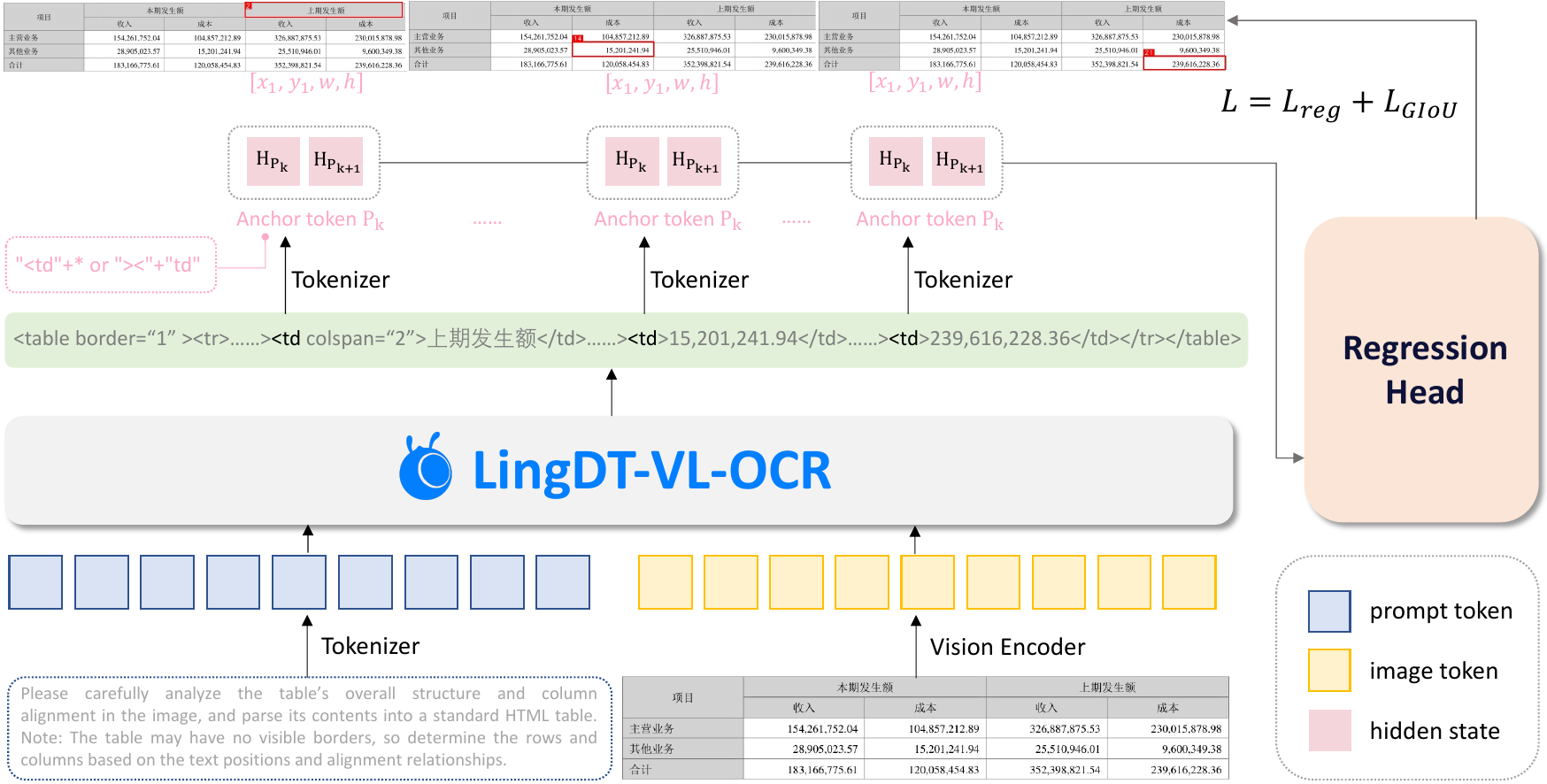}
    \caption{\textbf{Architecture of CellBBoxRegressor}. CellBBoxRegressor grounds each table cell by regressing its bounding box from decoder hidden states anchored at cell-start tokens.}
    \label{fig:regressor_vis}
\end{figure}

Financial document understanding in real-world business settings, such as auditing and compliance, often requires fine-grained cell-level visual references. However, most existing table parsing models focus on recovering the table structure (e.g., HTML) and do not provide reliable cell-level grounding that maps each predicted cell back to its precise location in the source document.
While modern VLMs~\cite{bai2025qwen3} have been explored for generic grounding tasks, a common remedy is to inject dedicated special tokens (e.g., \texttt{<bbox>}) ~\cite{lu2025bounding} and train the model to emit coordinates.
However, introducing extra special tokens increases sequence length and decoding burden, especially for tables with enormous cells, and may require non-trivial changes to the vocabulary and training recipe.
To meet the cell-level grounding requirement for document tables, we propose CellBBoxRegressor, as shown in Figure~\ref{fig:regressor_vis}, which leverages structural anchor tokens in the HTML stream to regress bounding boxes for each cell without introducing additional special tokens.

\subsubsection{Structural anchor tokens.}
To enable cell-level grounding with minimal changes to the generation pipeline, we avoid introducing an extra \texttt{<bbox>} token and instead reuse cell-start anchor tokens from the HTML stream. Due to different HTML serialization styles and the tokenization, the beginning
of a cell start tag \texttt{<td ...>} can be detected in two ways: (i) a start-tag token
\texttt{``<td''}; or (ii) an adjacent tag-boundary pair \texttt{``><''}+\texttt{``td''} in compact HTML.
We unify both cases by defining an anchor index $P_k$ that is aligned to the start of the $k$-th cell
start tag:
\begin{equation}
P_k \in \{\,t \mid y_t=\texttt{``<td''}\,\}\ \cup\ \{\,t+1 \mid (y_t=\texttt{``><''} \wedge y_{t+1}=\texttt{``td''})\,\}.
\end{equation}
Given the HTML token sequence $Y=\{y_t\}_{t=1}^{T}$ and decoder hidden states
$\mathbf{H}=\{\mathbf{H}_t\}_{t=1}^{T}$ with $\mathbf{H}_t\in\mathbb{R}^{D}$, we represent each cell
using a fixed two-token anchor window:
\begin{equation}
\mathbf{s}_k=\mathrm{Pool}\big(\mathbf{H}_{P_k},\mathbf{H}_{P_k+1}\big)\in\mathbb{R}^{D_s}.
\end{equation}
Here $\mathbf{H}_{P_k+1}$ corresponds to the immediate next token after the anchor (e.g., \texttt{``>''} or
an attribute-related token), providing local context while keeping a consistent representation across
cells.

\subsubsection{Training and inference.}
Training uses teacher forcing so the anchor indices $P_k$ appear deterministically in $Y$, enabling
one-to-one alignment between $\mathbf{s}_k$ and $\mathbf{b}_k$. We optimize a box loss (e.g., $\ell_1$
with optional GIoU):
\begin{equation}
\mathcal{L}_{\text{box}}=\frac{1}{M}\sum_{k=1}^{M}\lVert \hat{\mathbf{b}}_k-\mathbf{b}_k\rVert_1
\;(+\lambda\,\mathcal{L}_{\text{giou}}).
\end{equation}

Each cell is supervised by a normalized box
\begin{equation}
\mathbf{b}_m=\big(x^{(m)}_1, y^{(m)}_1, w^{(m)}, h^{(m)}\big), \quad \mathbf{b}_m\in[0,1]^4,
\end{equation}
where $(x_1,y_1)$ is the top-left corner and $(w,h)$ is width/height (normalized by $(W,H)$).
, where $(x_1,y_1)$
is the top-left corner and $(w,h)$ is width/height. A lightweight regressor predicts
\begin{equation}
\hat{\mathbf{b}}_m=g_\theta(\mathbf{s}_m),\quad g_\theta:\mathbb{R}^{D_s}\rightarrow \mathbb{R}^{4},\qquad
\hat{\mathbf{B}}=g_\theta(\mathbf{S})\in\mathbb{R}^{M\times 4}.
\end{equation}
For GIoU, we convert to corners
$\hat{\mathbf{b}}^{xyxy}_m=(\hat{x}_1,\hat{y}_1,\hat{x}_1+\hat{w},\hat{y}_1+\hat{h})$ and clamp to $[0,1]$.

During inference, we locate all anchor indices $P_k$ in the generated HTML using the same rule as in
training, extract $\mathbf{H}_{P_k}$ and $\mathbf{H}_{P_k+1}$ to form $\mathbf{s}_k$, and apply $g_\theta$
to obtain $\hat{\mathbf{B}}$, achieving cell-level localization without external detectors.

\subsection{FinDocBench}
\begin{figure*}[htbp]
    \begin{center}
        \includegraphics[width=1.0 \linewidth]{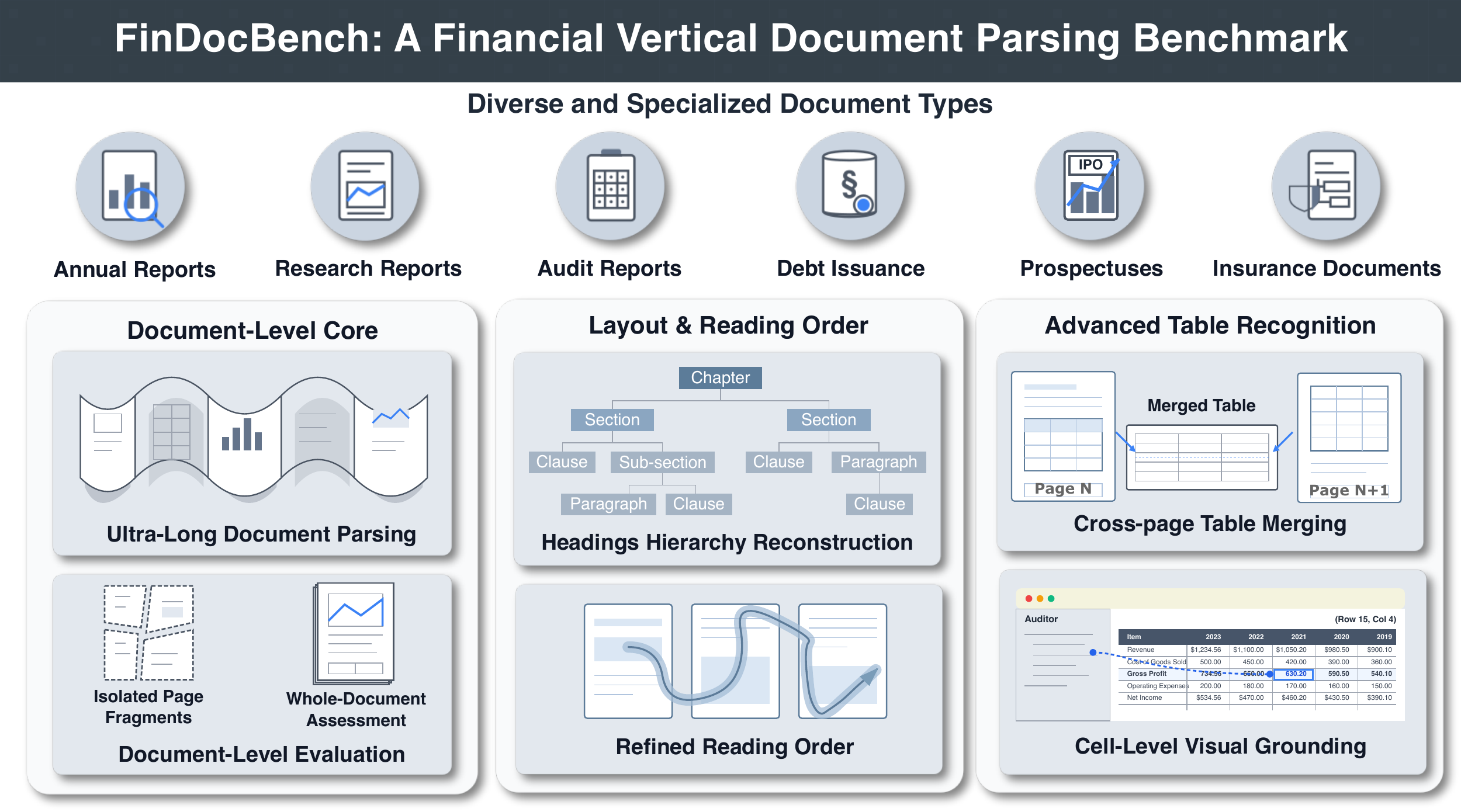}
    \end{center}
    \vspace{-0.2cm}
    \caption{ \textbf{Overview of FinDocBench.} It contains six financial document categories and provides a comprehensive evaluation focusing on ultra-long document parsing, hierarchical heading reconstruction, and advanced table recognition.}
    \label{fig:FinDocBench}
    \vspace{-0.2cm}
\end{figure*}

While OmniDocBench v1.5~\cite{ouyang2025omnidocbench} offers comprehensive evaluations across various general document parsing sub-tasks, the financial vertical imposes following distinct requirements:
\begin{enumerate}
  \item financial documents are characterized by \textbf{extreme length}, e.g., prospectuses and annual reports typically encompass hundreds of pages;
  \item the inherent complexity of financial vertical manifests in \textbf{hierarchical headings}, prevalent \textbf{cross-page content}, \textbf{complex layouts} and \textbf{intricate tables};
  \item the auditing demands in financial scenarios necessitate \textbf{precise cell-level localization in tables}.
\end{enumerate}
Existing benchmarks only sporadically address these critical features. To bridge these gaps, we introduce FinDocBench, a benchmark specifically designed to evaluate document parsing performance within the financial vertical. The overview of key features of FinDocBench is illustrated in Figure~\ref{fig:FinDocBench}.


\subsubsection{Data Collection}
\begin{figure*}[b]
    \begin{center}
        \includegraphics[width=1.0 \linewidth]{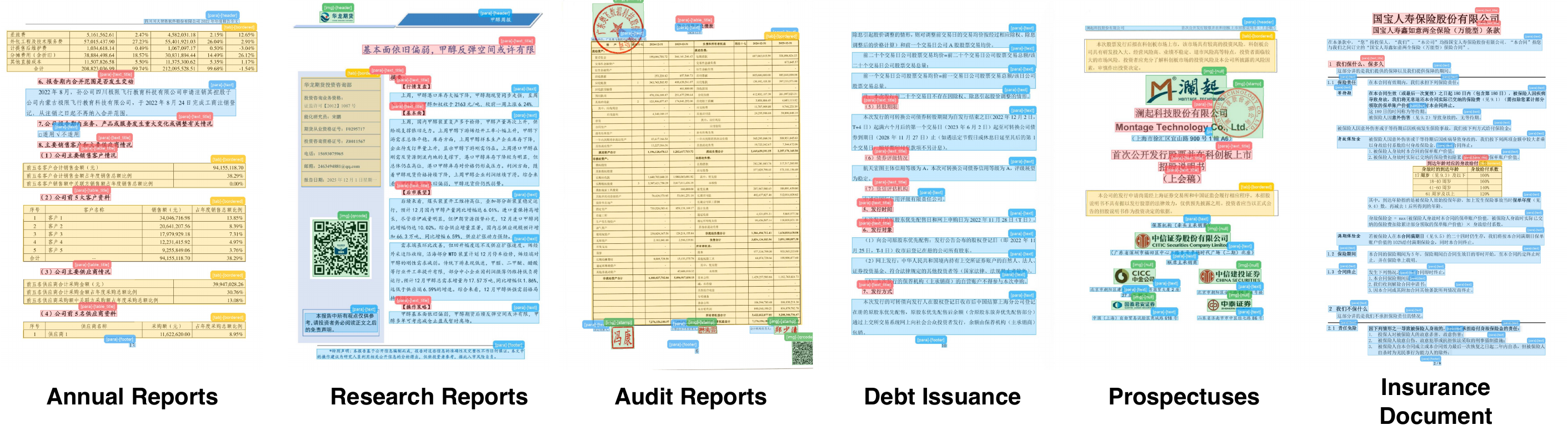}
    \end{center}
    \vspace{-0.2cm}
    \caption{\textbf{Typical cases of each financial document sub-category.}}
    \label{fig:document_examples}
    \vspace{-0.2cm}
\end{figure*}

We collect publicly available financial documents via search engines and various financial consulting websites. To ensure a comprehensive distribution, the benchmark encompasses six distinct sub-categories: annual reports, research reports, audit reports, debt issuance announcements, prospectuses, and insurance documents. Their individual examples are visualized in Figure~\ref{fig:document_examples}.

\subsubsection{Data Annotation}
\noindent
\textbf{Pre-annotation Phase}: Given that state-of-the-art (SOTA) closed-source commercial pipelines and open-source multimodal models have demonstrated robust performance in single-page parsing, we adopted a ensembling pre-annotation strategy.

\noindent
\textbf{Manual Correction Phase}: Beyond refining the pre-annotations, we put more emphasis on annotations of heading hierarchies in long documents, reading order with complex financial layouts, and intricate cross-page tables. Due to the parsing challenges in financial documents, current models still struggle with such complex structural content recovery, which necessitates meticulous human intervention to ensure the accuracy of annotations.

\noindent
\textbf{Expert Quality Assurance}: We further engaged Chartered Financial Analysts (CFAs) to perform data refinement, specifically focusing on the reading order of documents with highly complex layouts. This step ensures that the annotations in FinDocBench strictly adhere to professional financial standards and industry conventions.

\subsubsection{Statistics of FinDocBench}
\begin{table}[htbp]
    \centering
    \caption{\textbf{Statistics of the table recognition, layout with reading order parts in FinDocBench.}}
    \label{tab:table_layout_stats}
    \begin{tabular}{ccccc|cccc}
    \toprule
    \multicolumn{5}{c|}{\textbf{Table Recognition}} & \multicolumn{4}{c}{\textbf{Layout with Reading Order}} \\ 
    \midrule
    \multirow{2}{*}{\textbf{\makecell{\# of single-page \\ tables}}} & \multicolumn{3}{c}{\textbf{\# of cross-page tables}} & \multirow{2}{*}{\textbf{Total}} & \multirow{2}{*}{\textbf{\makecell{\# of layout \\ category}}} & \multicolumn{3}{c}{\textbf{\# of bboxes}} \\ 
    \cmidrule(lr){2-4} \cmidrule(l){7-9}
     & 2 & 3 & Above &  &  & Text & Others & Total \\ 
    \midrule
    572 & 425 & 28 & 19 & 1,044 & 17 & 18,349 & 20,284 & 38,633 \\ 
    \bottomrule
    \end{tabular}
\end{table}

\begin{figure*}[htbp]
    \begin{center}
        \includegraphics[width=1.0 \linewidth]{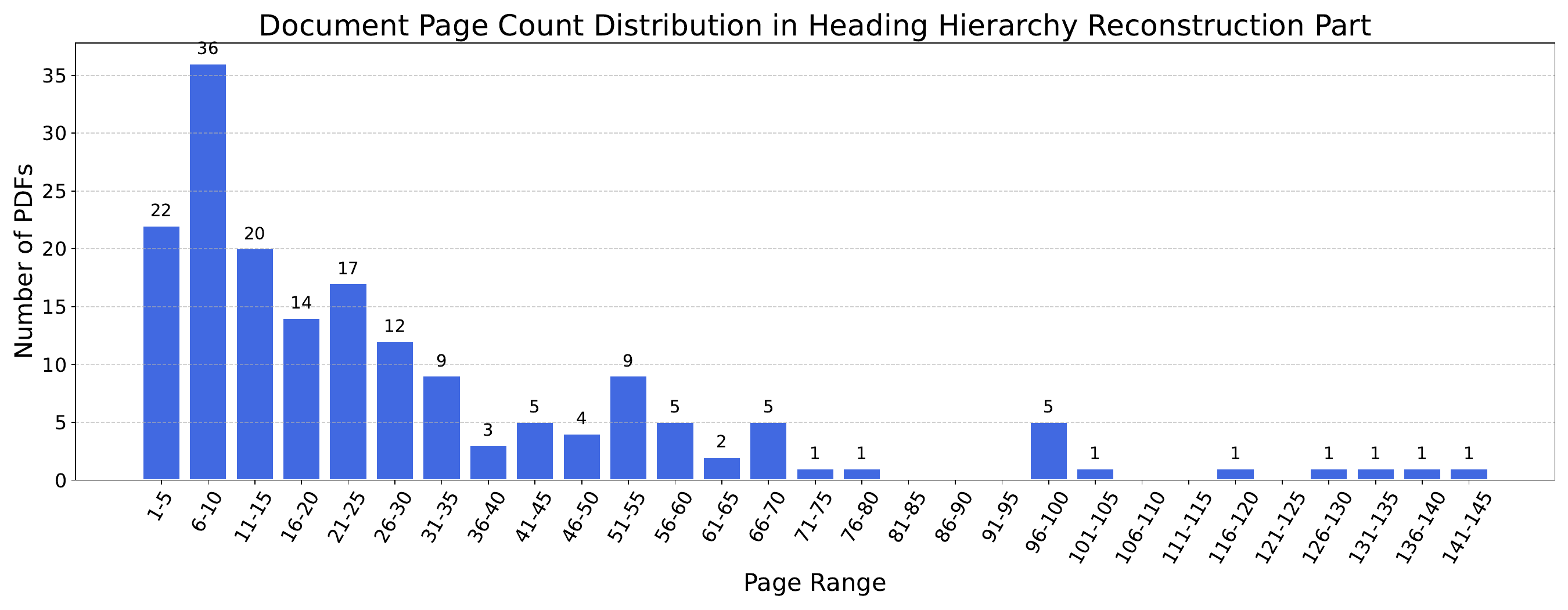}
    \end{center}
    \vspace{-0.2cm}
    \caption{ \textbf{Page count distribution in the heading hierarchy reconstruction part.} }
    \label{fig:page_count_distribution}
    \vspace{-0.5cm}
\end{figure*}

Following the data collection and annotation phases, the dataset is partitioned into three parts: table recognition, layout with reading order and heading hierarchy reconstruction. The corresponding statistical details are summarized in Table~\ref{tab:table_layout_stats} and Table~\ref{tab:heading_stats}. In the table recognition part, we intentionally emphasized cross-page tables due to their prevalence in financial scenarios. This part comprises a total of 1,044 tables, 472 of which span at least two pages. Regarding layout detection, we categorize document elements into 17 classes, with a total of 38,633 annotated instances with layout categories, reading order and contents. This part can serve as the benchmark for layout detection, reading order prediction and OCR simultaneously. For heading hierarchy reconstruction, we labeled 12,467 headings from 176 financial documents; these annotations contain the textual contents, hierarchical levels and spatial positions. The distribution of document lengths is illustrated in Figure~\ref{fig:page_count_distribution}, where 19.3\% of the documents are long-form, exceeding 30 or even 100 pages. We also calculated the average heading depth for various document types. Specifically, annual reports, audit reports, and prospectuses all reach the average depth of 3, reflecting the inherent structural complexity of these three document categories.

\begin{table}[htbp]
    \centering
    \caption{\textbf{Statistics of the heading hierarchy part in FinDocBench.}}
    \label{tab:heading_stats}
    \begin{tabular}{lcccc}
        \toprule
        \textbf{Document Type} & \textbf{\# of Docs} & \textbf{\# of Pages} & \textbf{\# of Headings} & \textbf{Avg Heading Depth} \\
        \midrule
        Annual Reports        & 27   & 1,328 & 4,331 & 3.32 \\
        Research Reports      & 67   & 842   & 1,257 & 1.63 \\
        Audit Reports         & 10   & 765   & 2,271 & 3.58 \\
        Debt Issuance         & 19   & 449   & 654   & 1.87 \\
        Prospectuses          & 25   & 1,409 & 2,850 & 3.52 \\
        Insurance Documents   & 28   & 286   & 1,104 & 2.10 \\
        \midrule
        \textbf{Total}        & \textbf{176} & \textbf{5,079} & \textbf{12,467} & \textbf{3.06} \\
        \bottomrule
    \end{tabular}
\end{table}

\subsection{Evaluation of FinDocBench}
\noindent
\textbf{Table of Contents edit-distance-based similarity (TocEDS).} The heading hierarchy of a document could be represented as an ordered multi-way tree. Based on the parsing results, we segment the document into fragments by recursively splitting according to heading levels. Within each fragment, headings of the same level are sorted by their reading order to reconstruct the Table of Contents as an ordered tree. TocEDS calculates the tree edit-distance-based similarity (TEDS) between the predicted tree and the ground truth, reflecting the model capability in recognizing heading contents and hierarchical levels.

\noindent
\textbf{Tree edit-distance-based similarity (TEDS) of concatenated cross-page table.} This evaluation extends TEDS to concatenated cross-page tables. It first prunes the table headers from non-initial pages and concatenates the fragments before calculating the TEDS metric against the ground truth. This metric provides an end-to-end assessment of table recognition accuracy across page boundaries.

\noindent
\textbf{Table Cell Intersection over Union (C-IoU).} This evaluation measures cell-level localization accuracy by computing the Intersection over Union (IoU) between the predicted and ground-truth bounding boxes for each table cell. Since general-purpose VLMs and OCR systems typically do not provide explicit cell grounding outputs, we report the cell-level grounding results of our CellBBoxRegressor; as shown in Table~\ref{tab:cell_iou}, it achieves strong localization performance, indicating accurate and audit-friendly cell-level referencing capability.

\section{Evaluation}

\subsection{OmniDocBench v1.5}
To establish a comprehensive baseline for general document parsing, we first evaluate our model on OmniDocBench v1.5~\cite{ouyang2025omnidocbench}, a widely recognized benchmark comprising 1,355 bilingual pages. Although our model is specifically tailored for financial scenarios, benchmarking on this public dataset allows for a standardized comparison with state-of-the-art methods. This evaluation serves as a foundation before assessing the model on our specialized financial benchmark.


\begin{table*}[htbp] 
\centering
\begin{threeparttable}
\caption{\textbf{Comparison of different models on document parsing tasks.} Best results are in \textbf{bold}.}
\label{tab:ocr_comparison}
\setlength{\tabcolsep}{1pt} 
\fontsize{7pt}{9pt}\selectfont

\begin{tabular}{l|l|c|ccccc}
\toprule
\textbf{Model Type} & \textbf{Methods} & 
\textbf{Overall$^{\uparrow}$} &
\textbf{Text$^{\text{Edit}}\downarrow$} & 
\textbf{Formula$^{\text{CDM}}\uparrow$} & \textbf{Table$^{\text{TEDS}}\uparrow$} & \textbf{Table$^{\text{TEDS-S}}\uparrow$} & \textbf{Reading Order$^{\text{Edit}}\downarrow$} \\ 
\midrule
\multirow{3}{*}{\textbf{Pipeline Tools}} 
& Marker-1.8.2~\cite{paruchuri2025marker} &71.30 & 0.206 & 76.66 & 57.88 & 71.17 & 0.250 \\
& MinerU-pipeline~\cite{wang2024mineru} &75.51 & 0.209 & 76.55 & 70.90 & 79.11 & 0.225 \\
& PP-StructureV3~\cite{cui2025paddleocr3} &86.73 & 0.073 & 85.79 & 81.68 & 89.48 & 0.073 \\ 
\midrule
\multirow{5}{*}{\textbf{General VLMs}} 
& InternVL3.5-241B~\cite{wang2025internvl3} & 82.67  & 0.142 & 87.23 & 75.00 & 81.28 & 0.125 \\
& Gemini-2.5 Pro~\cite{team2023gemini}& 88.03  & 0.075 & 85.82 & 85.71 & 90.29 & 0.097 \\ 
& Qwen3-VL-235B-Instruct~\cite{bai2025qwen3} & 89.15  & 0.069 & 88.14 & 86.21 & 90.55 & 0.068\\ 
\midrule
\multirow{12}{*}{\textbf{Specialized VLMs}} 
& Dolphin~\cite{feng2025dolphin} & 74.67  & 0.125 & 67.85 & 68.70 & 77.77 & 0.124 \\
& OCRFlux-3B~\cite{chatdoccom2025ocrflux} & 74.82  & 0.193 & 68.03 & 75.75 & 80.23 & 0.202 \\
& Mistral OCR~\cite{mistralai2025mistralocr} & 78.83  & 0.164 & 82.84 & 70.03 & 78.04 & 0.144 \\
& POINTS-Reader~\cite{liu2025points}& 80.98  & 0.134 & 79.20 & 77.13 & 81.66 & 0.145 \\
& olmOCR-7B~\cite{poznanski2025olmocr} & 81.79  & 0.096 & 86.04 & 68.92 & 74.77 & 0.121 \\
& MinerU2-VLM~\cite{wang2024mineru} & 85.56 & 0.078 & 80.95 & 83.54 & 87.66 & 0.086 \\
& Nanonets-OCR-s~\cite{mandal2025nanonets}&85.59  & 0.093 &85.90 & 80.14 & 85.57 & 0.108 \\
& MonkeyOCR-pro-1.2B~\cite{li2025monkeyocr}& 86.96  & 0.084 & 85.02 & 84.24 & 89.02 & 0.130 \\
& MonkeyOCR-3B~\cite{li2025monkeyocr} & 87.13 & 0.075 & 87.45 & 81.39 & 85.92 & 0.129 \\
& dots.ocr~\cite{li2025dots} & 88.41  & 0.048 & 83.22 & 86.78 & 90.62 & 0.053 \\
& MonkeyOCR-pro-3B~\cite{li2025monkeyocr} & 88.85  & 0.075 & 87.25 & 86.78 & 90.63 & 0.128 \\
& MinerU2.5~\cite{niu2025mineru2} & 90.67  & 0.047 & 88.46 & 88.22 & 92.38 & 0.044 \\
& DeepSeek-OCR2~\cite{wei2026deepseek}  & 91.09  & 0.048 & 90.31  &  87.75 & 92.06 & 0.057 \\
\rowcolor{blue!5}
& PaddleOCR-VL~\cite{cui2025paddleocr} & 92.86  & 0.035 & 91.22 & 90.89 & 94.76 & 0.043 \\ 
\rowcolor{blue!5}
& GLM-OCR~\cite{glmocr2026} & 93.94   & 0.043 & 93.51 & 92.60 & 94.84 & 0.043 \\
\rowcolor{blue!5}
& PaddleOCR-VL-1.5~\cite{cui2026paddleocr1_5} & \textbf{94.50}  & 0.035 & \textbf{94.21} & \textbf{92.76} & \textbf{95.79} & \textbf{0.042} \\
\rowcolor{blue!5}
& Ours & 92.91 & \textbf{0.034} & 90.78 & 91.34 & 93.85 & 0.044 \\ 
\bottomrule
\end{tabular}

\begin{tablenotes}
    \footnotesize
    \item The formula of Overall is calculated as: ((1 - Text$^{Edit}$) $\times$ 100 +Formula$^{CDM}$+Table$^{TEDS}$) / 3.
    
\end{tablenotes}
\end{threeparttable}
\end{table*}




As illustrated in Table \ref{tab:ocr_comparison}, our model achieves state-of-the-art results in text recognition, securing the best Text$^{\text{Edit}}$ score of 0.034, outperforming other high-performing models such as PaddleOCR-VL-1.5 (0.035). This precision in character-level recognition ensures high fidelity in extracting critical financial terminology and figures.

In terms of table parsing, our model maintains a significant lead over prominent specialized models, including MinerU2.5~\cite{niu2025mineru2} and DeepSeek-OCR2~\cite{wei2026deepseek}, with a Table$^{\text{TEDS}}$ score of 91.34 and a Table$^{\text{TEDS-S}}$ 
score of 93.85. While PaddleOCR-VL-1.5 shows a slight edge in structural metrics, our model's performance remains highly robust. Given that financial reports often contain dense and structurally intricate tables, this superior performance in TEDS metrics compared to most industry-standard models validates its effectiveness in recovering complex cell relationships.

Overall, the evaluation on OmniDocBench v1.5~\cite{ouyang2025omnidocbench} confirms that our model possesses top-tier general document parsing capabilities, achieving an Overall score of 92.91. Its exceptional accuracy, particularly the SOTA text recognition and reliable table structure recovery, provides a solid foundation for downstream financial data analysis.


\subsection{Layout Detection \& Reading Order Results}

To evaluate the performance of our framework on complex financial documents, we conducted comparative experiments against the general-purpose PP-DocLayoutV3 baseline. Our layout analysis module is built upon the PP-DocLayout-plus-L architecture, which was further fine-tuned on a domain-specific dataset comprising insurance policies, financial reports, and bank statements. This specialized training allows the model to better capture the intricate structures (e.g., nested tables and dense multi-column text) common in financial scenarios.

Furthermore, to address reading order prediction, we implemented a 6-layer Transformer-based module that incorporates the spatial-relation modeling approach from PaddleOCR-VL. This architecture leverages self-attention mechanisms to effectively capture global dependencies among disparate layout elements. As demonstrated in Table \ref{tab:financial layout analysis}, our proposed model achieves superior performance on FinDocBench, reaching a 0.873 mAP@0.5:0.95, a 0.112 improvement over PP-DocLayoutV3. More notably, our model significantly reduces the Average Relative Distance (ARD) from 0.443 to 0.075. These results validate the necessity of domain-specific fine-tuning and robust sequence modeling for maintaining logical coherence in financial document parsing. Qualitative results provided in Appendix~\ref{app:layout_analysis} and Appendix~\ref{app:layout_analysis_bad_cases} further illustrate the model's precision in layout detection and reading order prediction.

\subsection{Document-level Heading Hierarchy Reconstruction} \label{sec:result-dhr}
Table~\ref{tab:document-level heading hierarchy} compares our method against
the text-only baseline across six financial document types, sorted by average
document length. Our method achieves a higher overall TocEDS score (0.6273 vs.
0.5643), with the largest gains on longer document types: +18.5\% on Audit
Reports (76.50p) and +7.4\% on IPO Prospectuses (56.36p). This confirms that the
pseudo-TOC's visual and spatial cues are particularly effective when heading
hierarchies span many pages, where text alone struggles to resolve level
ambiguities.
On shorter documents, however, the two methods perform comparably, with
Text-Only marginally outperforming ours on Research Reports (12.57p) and
Insurance Documents (10.21p).
This is expected, as shorter documents tend to have simpler and less ambiguous
heading structures that are already well-captured by text alone.

\subsection{Table Parsing Results}



\paragraph{Attribute Correlation with Parsing Quality.}
We analyze how each attribute in our tabular label schema correlates with parsing quality. Table~\ref{tab:attr_corr} reports the Pearson correlation between attribute values and TEDS on our in-house set. Overall, structural complexity (e.g., larger rowspan/colspan) and ICD show the strongest negative correlations with TEDS, making them effective signals for selecting high quality training data.

\begin{table}[t]
    \centering
    \caption{\textbf{Financial Layout Analysis}}
    \label{tab:financial layout analysis}
    \begin{tabular}{lcc} 
        \toprule
        Method & mAP$^{\text{0.5:0.95}}$ $\uparrow$ & Reading Order (ARD) $\downarrow$ \\ 
        \midrule
        PP-DocLayoutV3 & 0.761 & 0.443 \\
        \midrule
        Ours  & \textbf{0.873} & \textbf{0.075} \\ 
        \bottomrule
    \end{tabular}
\end{table}

\begin{table}[t]
    \centering
    \caption{
        \textbf{Effect of the pseudo-TOC on document-level heading hierarchy
        reconstruction.}
        Numbers in the header row (e.g., 76.50p) denote the average document
        length in pages.
    }
    \label{tab:document-level heading hierarchy}
    \scalebox{.92}{\begin{tabular}{@{}l@{}ccccccc@{}}
        \toprule
        Method & \makecell{Overall\\\small{(avg. pages)}} & \makecell{Audit\\Reports\\\small{(76.50p)}} & \makecell{IPO\\Prospectuses\\\small{(56.36p)}} & \makecell{Annual\\Reports\\\small{(49.19p)}} & \makecell{Debt\\Issuance\\\small{(23.63p)}} & \makecell{Research\\Reports\\\small{(12.57p)}} & \makecell{Insurance\\Documents\\\small{(10.21p)}} \\
        \midrule
        Text-Only & 0.5643 & 0.4322 & 0.7641 & 0.5290 & 0.7448 & \textbf{0.3808} & \textbf{0.3539} \\
        \textbf{Text+Image (Ours)} & \textbf{0.6273} & \textbf{0.6168} & \textbf{0.8385} & \textbf{0.5893} & \textbf{0.7457} & 0.3749 & 0.3487 \\
        \bottomrule
    \end{tabular}}
\end{table}

\begin{table}[t]
\centering
\caption{\textbf{Pearson correlation between table attributes and TEDS.} Abbreviations: Empty Rat.=empty-cell ratio; RS=count of rowspan; CS=count of colspan; Max RS/Max CS=maximum value of rowspan/colspan; Diff. Std/Diff. Range are defined as the standard deviation/range of TEDS across repeated high-temperature inference runs for a given sample.}
\label{tab:attr_corr}
\begin{tabular}{cccccccc}
\toprule
Attribute & Empty Rat. & Max RS & RS & Max CS & CS & Diff. Std & Diff. Range \\
\midrule
Correlation & -0.027 & -0.116 & -0.174 & -0.283 & -0.318 & -0.324 & -0.332 \\
\bottomrule
\end{tabular}
\end{table}

\paragraph{Quantitative results.}
As shown in Table 7, our approach outperforms or matches representative SOTA methods like MinerU2.5~\cite{niu2025mineru2} and PaddleOCR-VL-1.5~\cite{cui2026paddleocr1_5} across diverse benchmarks. Notably, when specifically evaluated on the isolated table subset of OmniDocBench v1.5, our model achieves the highest TEDS score of 93.2, surpassing all other models.

It is worth noting that this score (93.2) is higher than the TEDS recorded in the full-page evaluation (91.34 in Table \ref{tab:ocr_comparison}). This discrepancy arises because the full-page "Overall" evaluation involves a complete Markdown prediction, where table metrics can be negatively impacted by upstream layout detection errors or noise in the text-to-ground-truth matching process. By isolating the table parsing task, we demonstrate that our core architecture possesses superior structural recovery capabilities. The consistent performance gains across both public and private datasets further validate the efficacy of the proposed Curriculum Learning and Reinforcement Optimization strategies in mastering complex tabular structures.

\paragraph{Cross-page Table Merging Results}
To evaluate the efficacy of the proposed cross-page merging mechanism, we curated a specialized dataset comprising 472 cross-page table instances. The experimental pipeline involved two stages: first, individual table fragments on each page were parsed using our VLM model ; second, the adaptive merging algorithm described in Section 3.2 was applied to unify these fragments into integrated entities. The results demonstrate that our approach achieves an average TEDS of \textbf{0.8915} across the entire test set. This high similarity score validates the robustness of the heuristic-based splicing logic in preserving the global structural integrity of complex, cross-page tables.


\begin{table}[t]
\centering
\caption{\textbf{TEDS comparison on different benchmarks.} $^{\ast}$denotes the Table subset, \dag denotes results measured by ourselves.}
\label{tab:teds_comparison_small}
\begin{tabular}{lcccc}
\toprule
\textbf{Method} & OmniDocBench v1.5$^{\ast}$ & PubTabNet & In-house & FinDocBench$^{\ast}$ \\
\midrule
MinerU2.5~\cite{niu2025mineru2}     & 88.2 & \textbf{89.1} & 88.7 & 94.2 \\
PaddleOCR-VL-1.5~\cite{cui2026paddleocr1_5}  & 92.8 & 84.6 & 80.7 & 90.5 \\
GLM-OCR~\cite{glmocr2026}\textsuperscript{\dag}     & 92.6 & 85.2 & 84.4 & 93.0 \\
Our           & \textbf{93.2} & 87.8 & \textbf{90.1} & \textbf{95.7} \\
\bottomrule
\end{tabular}
\end{table}

\begin{table}[t]
    \centering
    \caption{\textbf{Cell-level grounding performance (C-IoU).}}
    \label{tab:cell_iou}
    \begin{tabular}{lccccc}
        \toprule
        Method & IoU@0.3 $\uparrow$ & IoU@0.5 $\uparrow$ & IoU@0.7 $\uparrow$ & Mean IoU $\uparrow$ & Median IoU $\uparrow$ \\
        \midrule
        Ours & 0.9765 & 0.9095 & 0.6411 & 0.7199 & 0.7555 \\
        \bottomrule
    \end{tabular}
\end{table}

\subsection{Table Cell-Level Visual Reference Results}

\paragraph{Quantitative results.}
We further report quantitative localization performance using Intersection-over-Union (IoU) between the predicted and ground-truth cell bounding boxes. In FinDocBench, we collect 572 tables from financial documents as the test set. Table~\ref{tab:cell_iou} summarizes the results on our evaluation set. The model achieves reasonable cell-level IoU, indicating that the proposed anchor-token-based regression learns a stable mapping from structural decoding states to spatial layout.

\begin{figure}[t]
    \centering
    \includegraphics[width=0.8\linewidth]{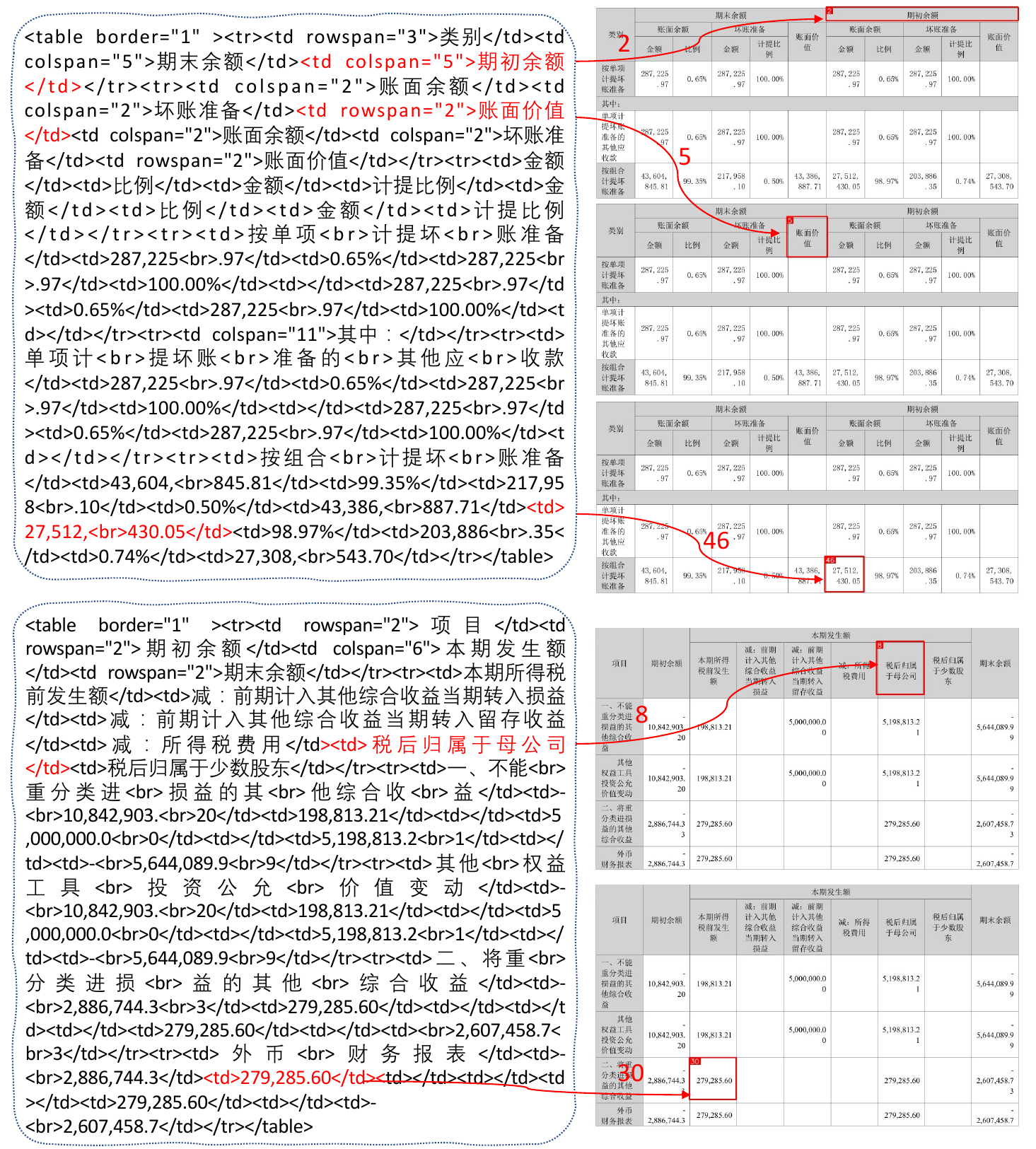}
    \caption{\textbf{Audit visualization of cell-level referencing.} Predicted cell boxes are overlaid on the input table images.}
    \label{fig:audit_vis}
\end{figure}

\paragraph{Visualization of Cell-Level Visual Reference.}
We visualize the predicted cell bounding boxes derived from structural anchor tokens and overlay them on the original table images. Figure~\ref{fig:audit_vis} shows representative examples. The visualizations indicate that our CellBBoxRegressor can localize most cells accurately, including merged cells and cells at arbitrary positions, providing reliable cell-level visual references for downstream auditing and provenance tracing.

\section{Conclusion}
In this work, we introduced LingDT-VL-OCR, a system designed to overcome semantic fragmentation in long financial documents by shifting from page-level to document-level parsing. Our framework ensures structural continuity through cross-page consolidation and Document-level Heading Hierarchy Reconstruction, which provides a consistent structural skeleton for high-performance downstream RAG application. To meet rigorous compliance standards, we achieved audit-grade table parsing by integrating curriculum learning with a novel CellBBoxRegressor. This mechanism leverages structural anchor tokens to enable cell-level visual referencing capability, allowing every data point to be mapped back to its source coordinates. Furthermore, we established FinDocBench, a specialized benchmark designed to evaluate the unique challenges of financial verticality, such as extreme length and complex layouts. Quantitative evaluations show that LingDT-VL-OCR achieves state-of-the-art results in table parsing while remaining highly competitive in general document parsing. By addressing the ``last-mile'' precision challenge, this system provides a robust foundation for automated financial workflows, with future work focusing on multilingual expansion and deeper agentic integration to further advance automated financial intelligence.


\bibliographystyle{plainnat}
\bibliography{main}
\newpage 
\appendix
\section*{Appendix}
\section{Cross-page table cases}
\label{app:Cross-page table cases}
\begin{figure}[H]
\centering
\begin{minipage}[b]{0.46\textwidth}
    \centering
    \includegraphics[width=\textwidth]{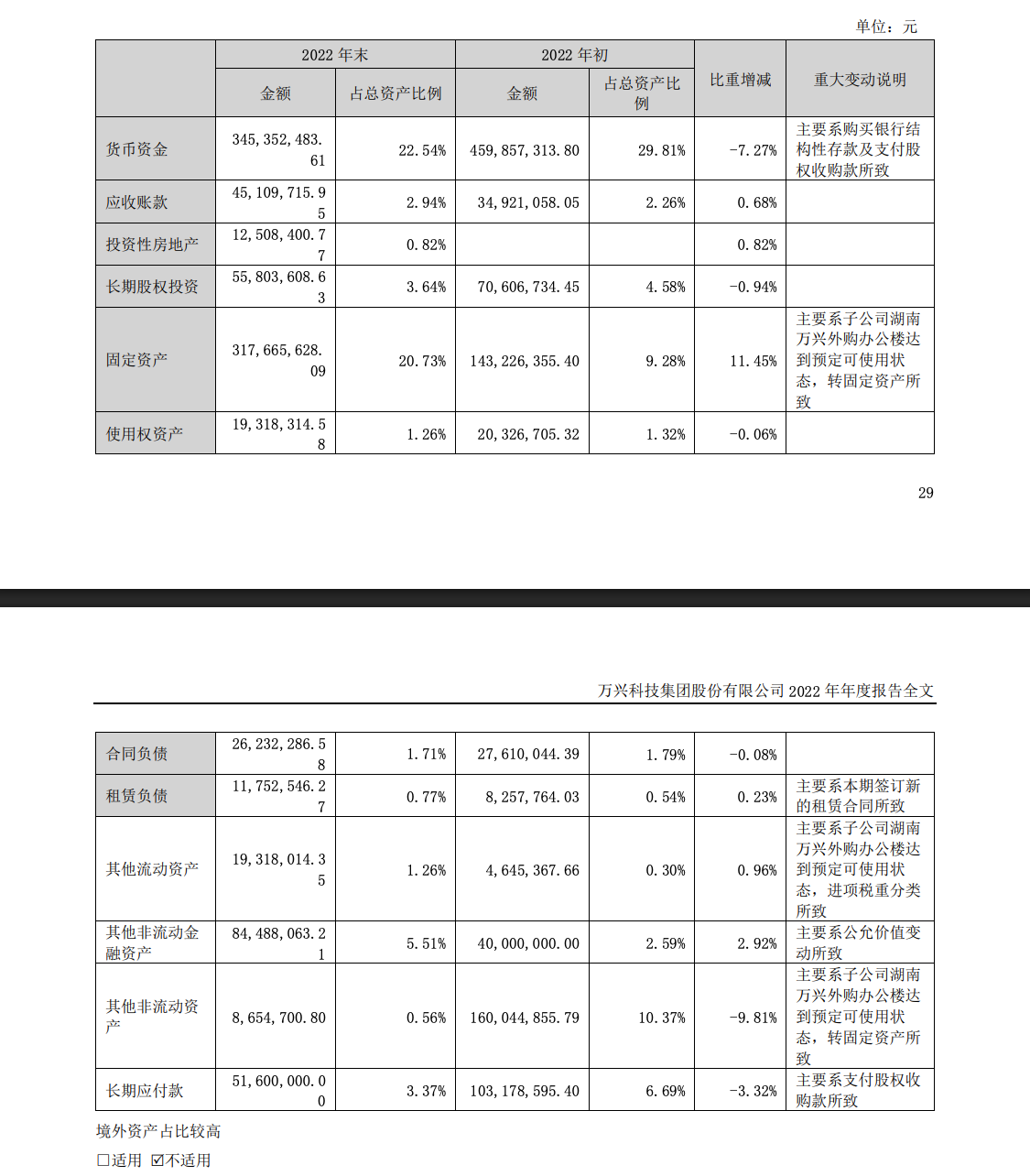}
    \vspace{2pt}
    \centerline{(a)}
\end{minipage}
\hspace{1cm} 
\begin{minipage}[b]{0.46\textwidth}
    \centering
    \includegraphics[width=\textwidth]{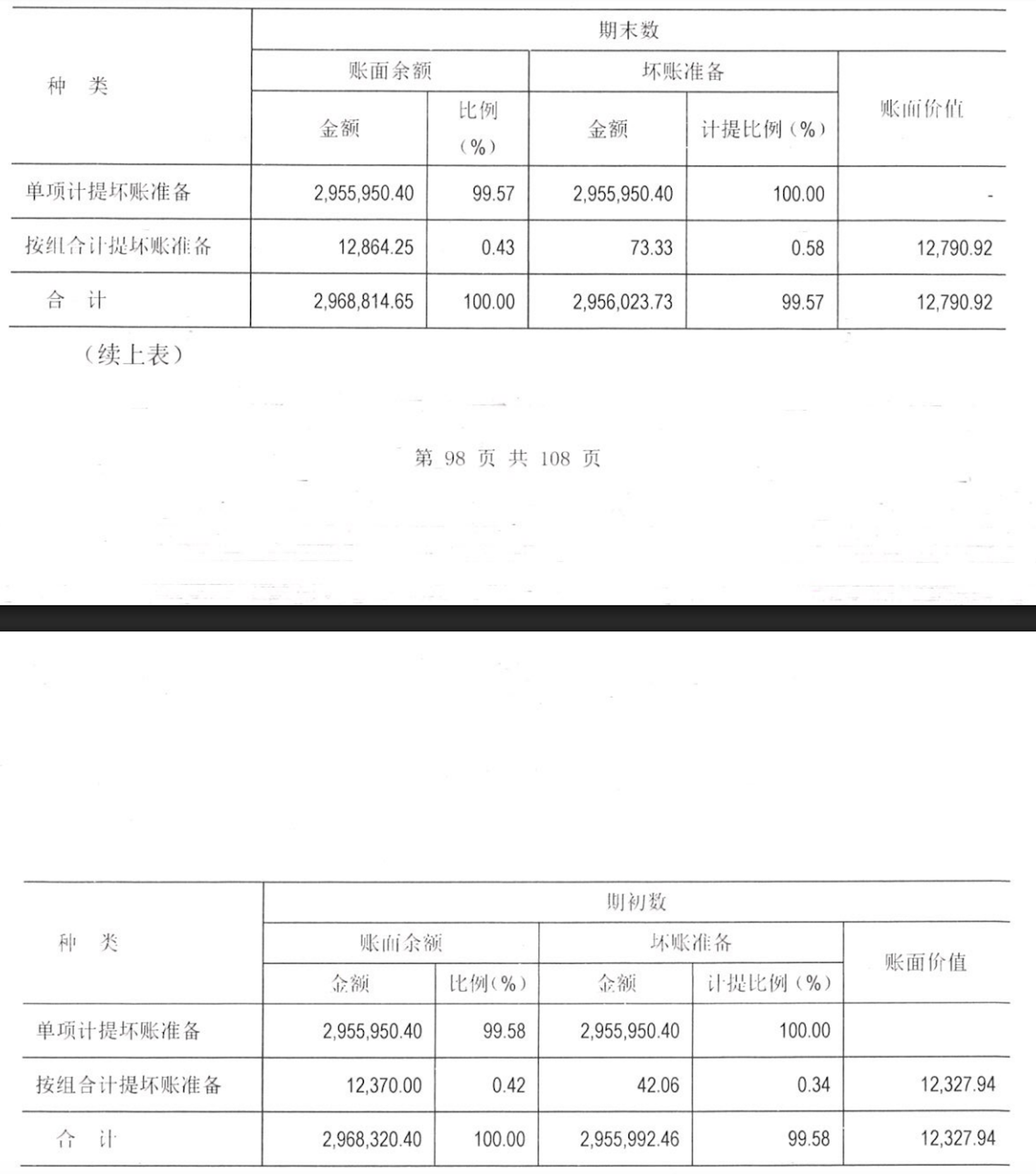}
    \vspace{2pt}
    \centerline{(b)}
\end{minipage}

\vspace{5pt}
\centerline{Homogeneous Splicing table} 

\vspace{15pt} 

\begin{minipage}[b]{0.46\textwidth} 
    \centering
    \includegraphics[width=\textwidth]{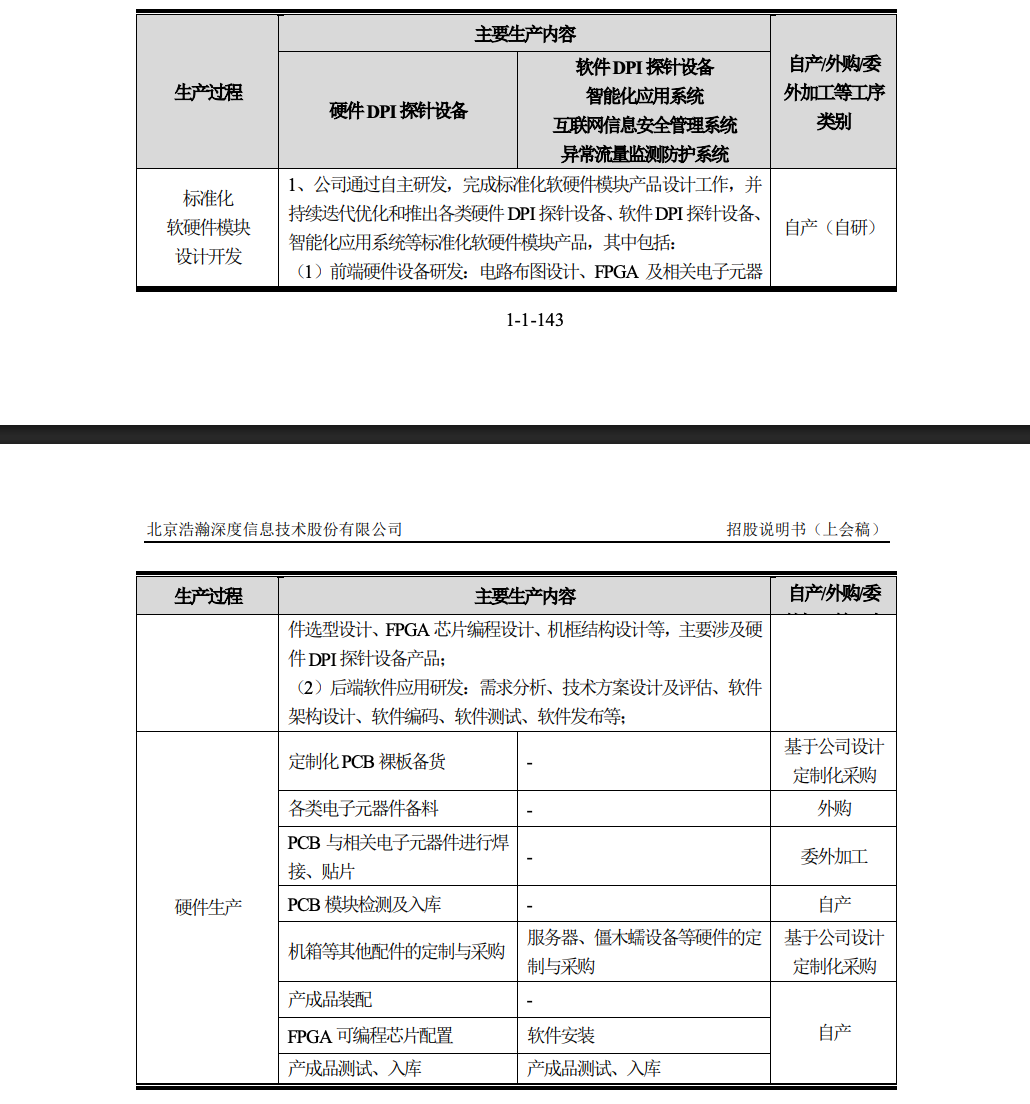}
    \vspace{2pt}
    \centerline{(c)}
    \vspace{5pt}
    \centerline{Heterogeneous Splicing table} 
\end{minipage}

\caption{\textbf{Visualization of cross-page tables.}}
\label{fig:appendix__figures}
\end{figure}





\section{Layout Analysis Results}
\label{app:layout_analysis}

\begin{figure}[H]
    \centering
    \begin{minipage}[b]{0.44\textwidth}  
        \centering
        \includegraphics[width=\textwidth]{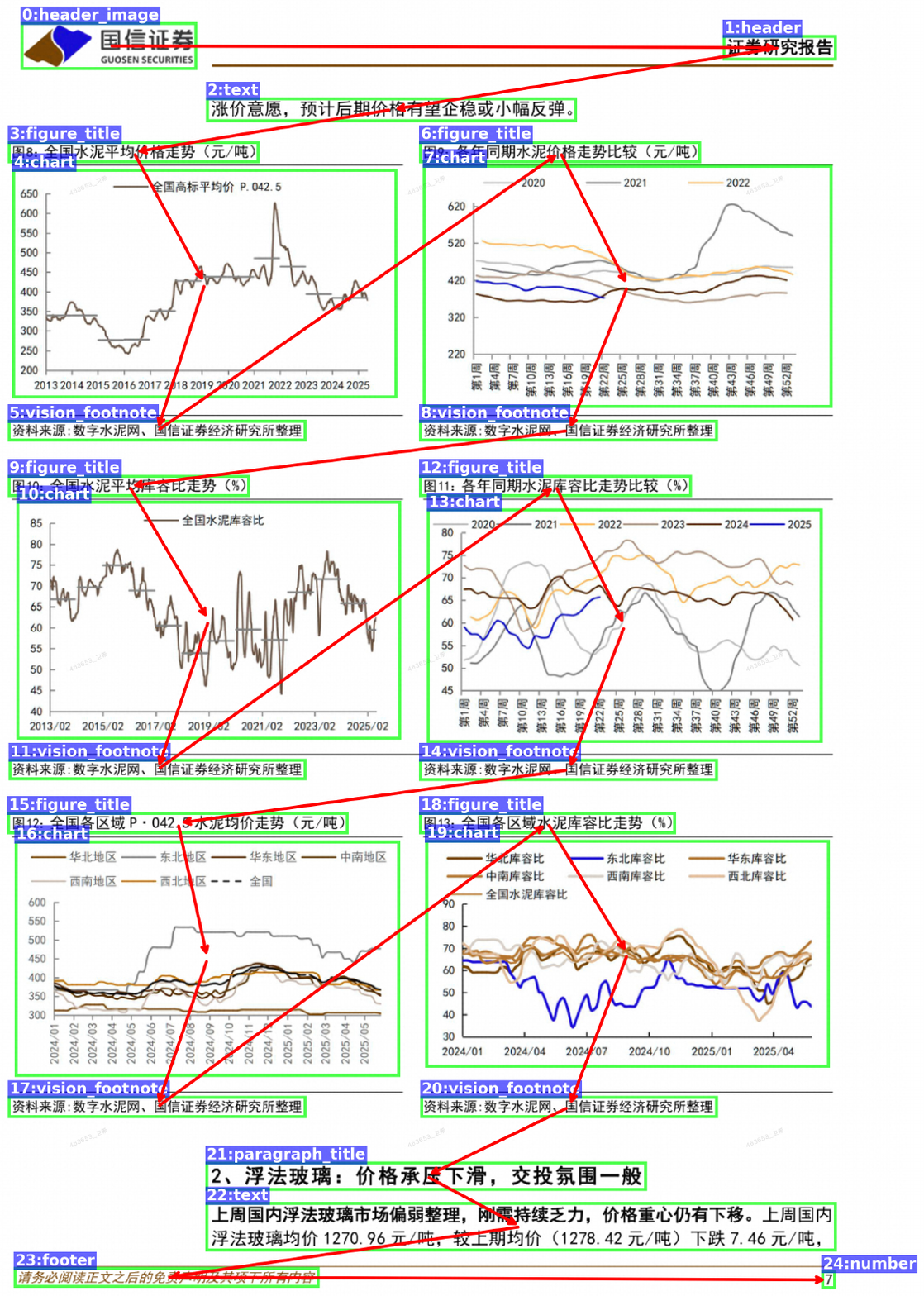}
        \vspace{2pt} 
        \centerline{(a)}
    \end{minipage}
    \hspace{1cm} 
    \begin{minipage}[b]{0.44\textwidth}
        \centering
        \includegraphics[width=\textwidth]{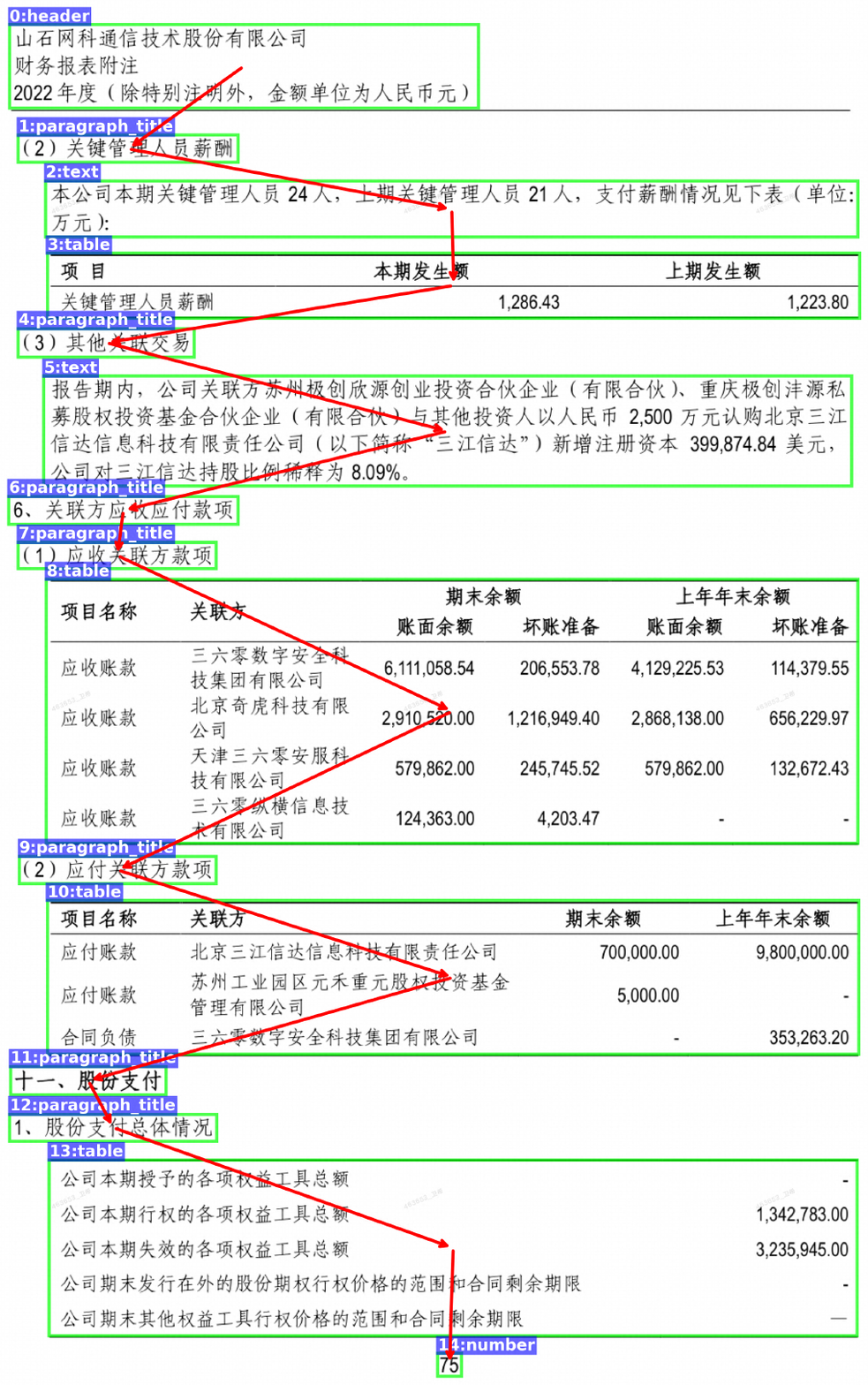}
        \vspace{2pt}
        \centerline{(b)}
    \end{minipage}

    \vspace{10pt} 

    \begin{minipage}[b]{0.44\textwidth}
        \centering
        \includegraphics[width=\textwidth]{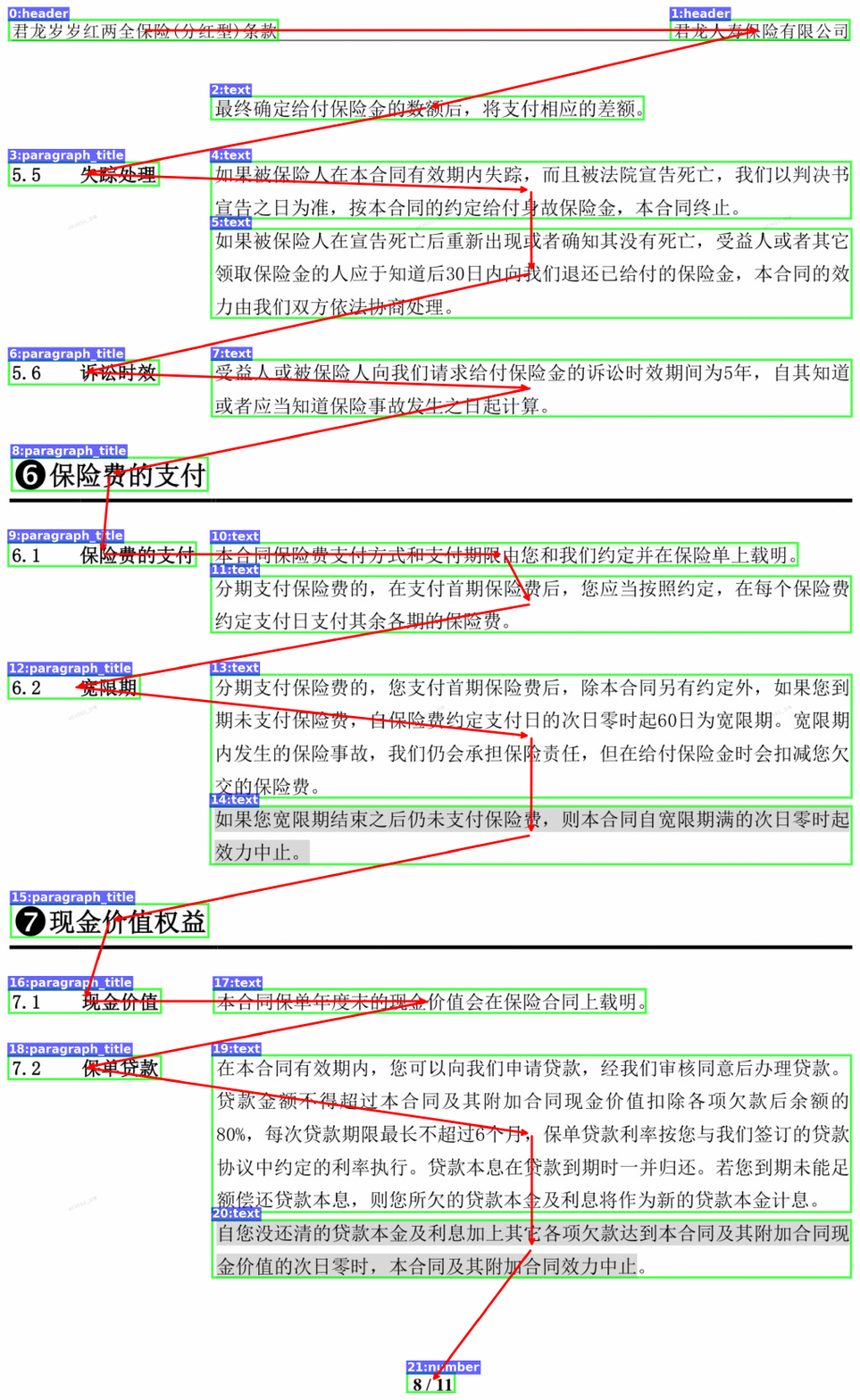}
        \vspace{2pt}
        \centerline{(c)}
    \end{minipage}
    \hspace{1cm}
    \begin{minipage}[b]{0.44\textwidth}
        \centering
        \includegraphics[width=\textwidth]{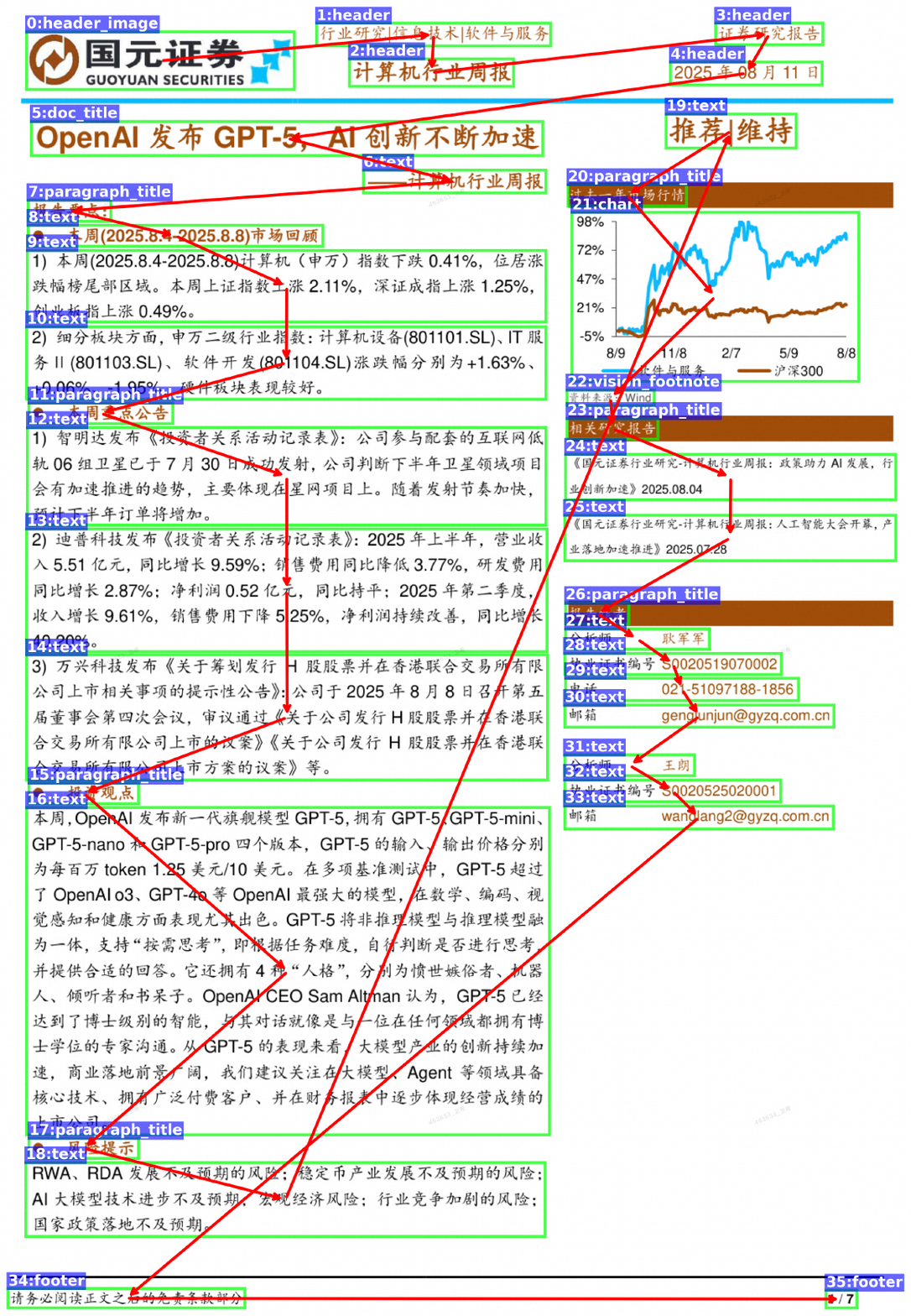}
        \vspace{2pt}
        \centerline{(d)}
    \end{minipage}

    \caption{\textbf{Visualization of layout analysis results.}}
    \label{fig:appendix_four_figures}
\end{figure}

\section{Layout analysis bad cases}
\label{app:layout_analysis_bad_cases}

\begin{figure}[H]
    \centering
    \begin{minipage}[b]{0.44\textwidth}
        \centering
        \includegraphics[width=\textwidth]{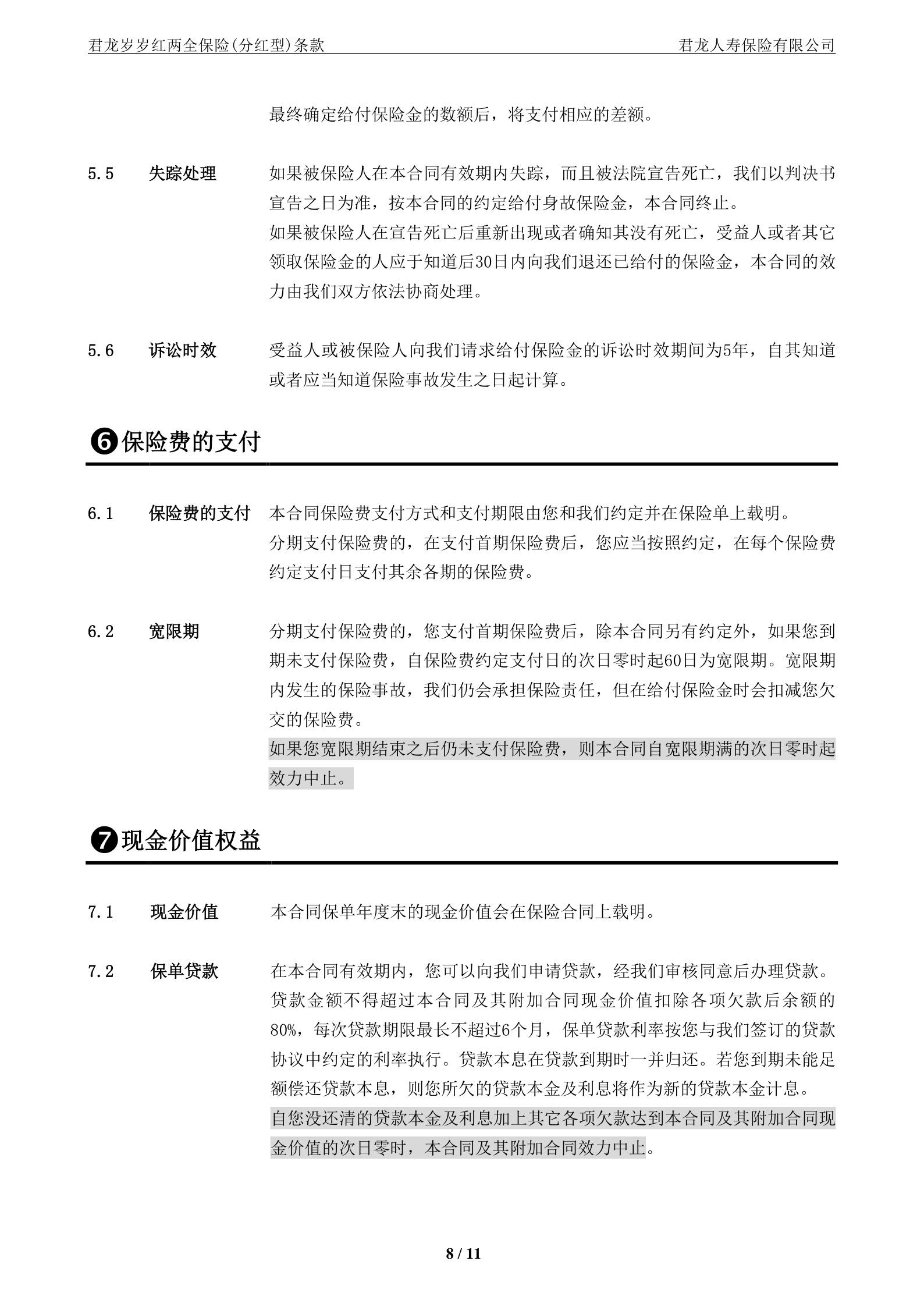}
        \centerline{(a)Raw image} 
    \end{minipage}
    \hfill
    \begin{minipage}[b]{0.44\textwidth}
        \centering
        \includegraphics[width=\textwidth]{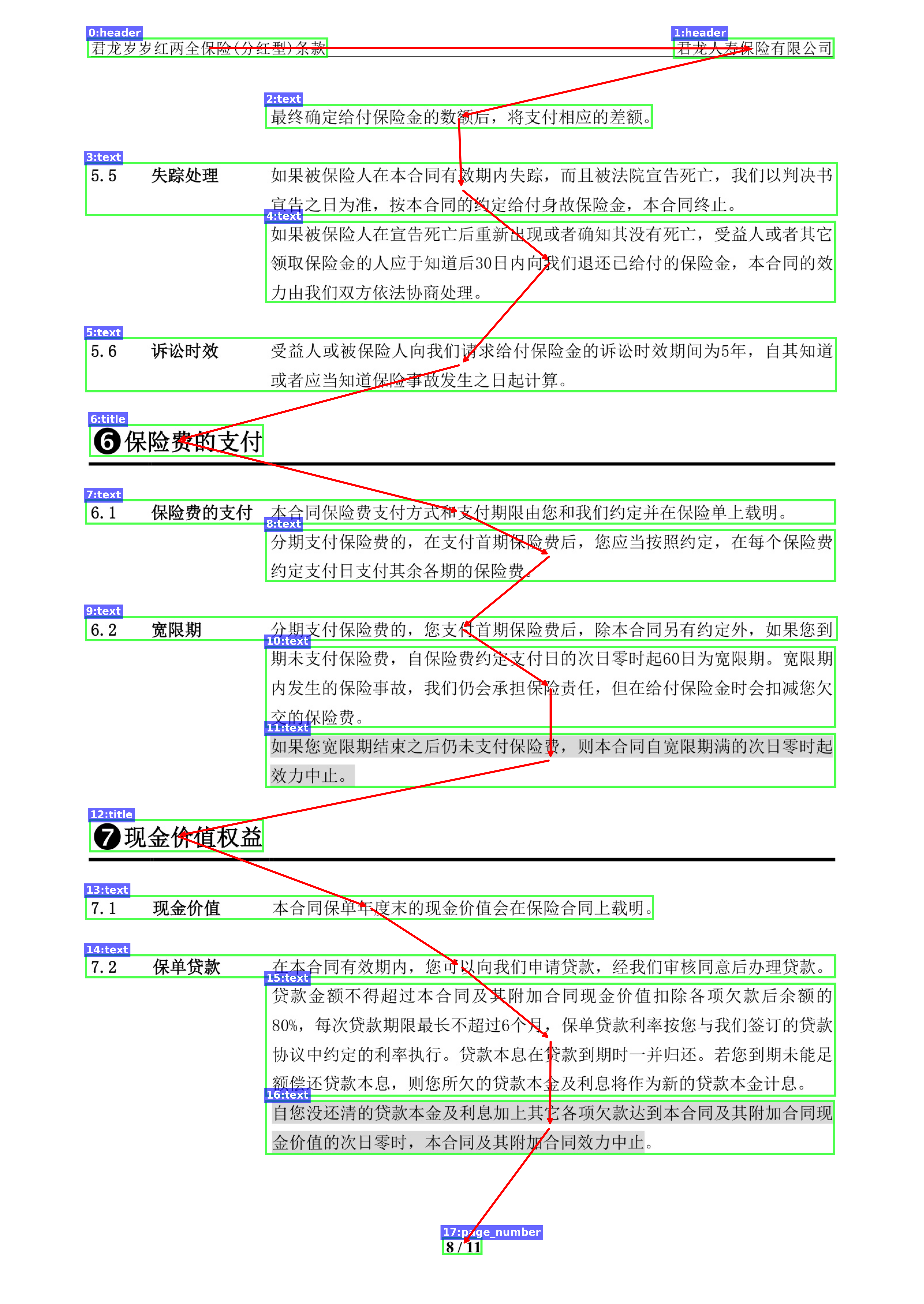}
        \centerline{(b)MinerU2.5} 
    \end{minipage}

    \vspace{15pt}

    \begin{minipage}[b]{0.44\textwidth}
        \centering
        \includegraphics[width=\textwidth]{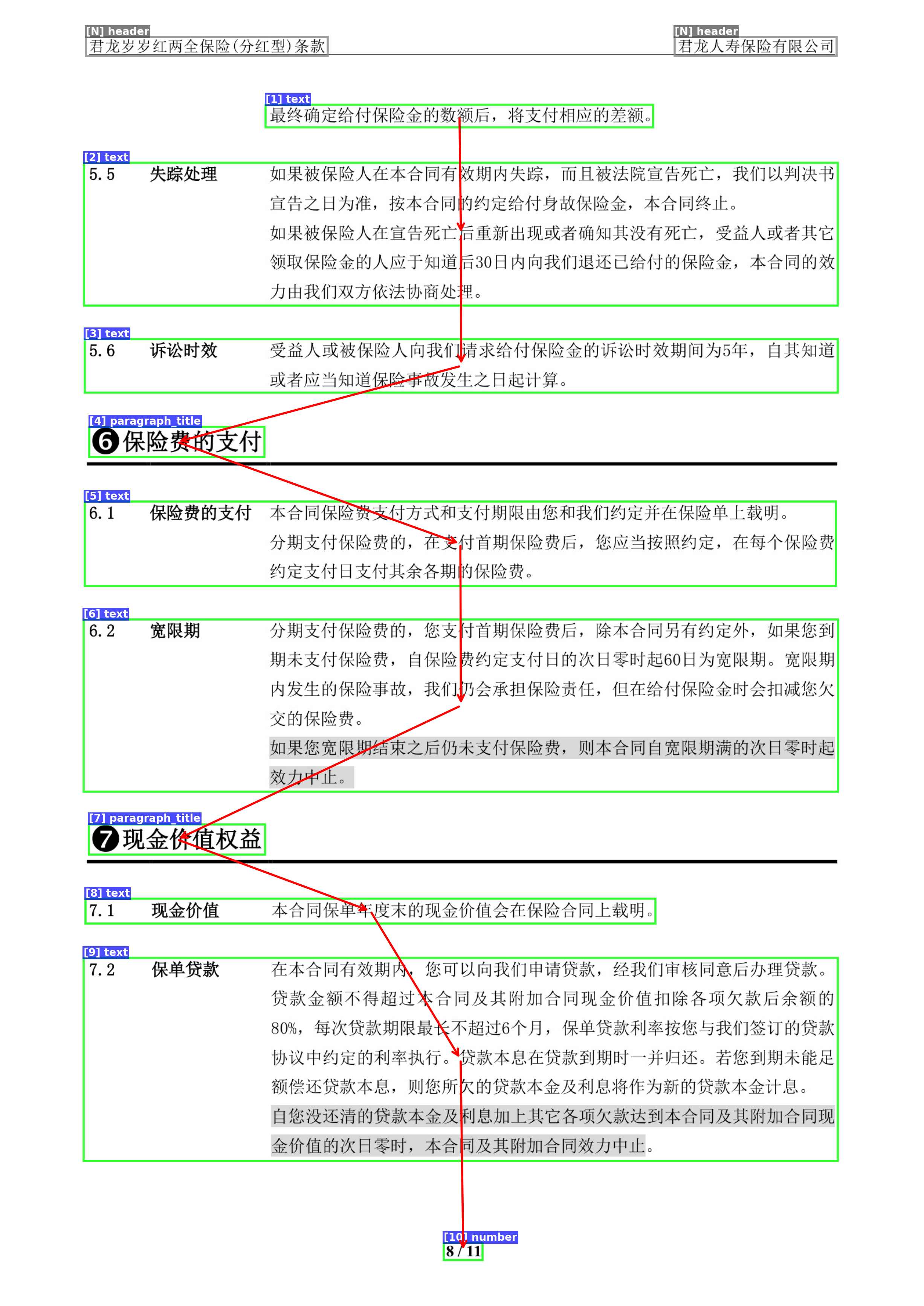}
        \centerline{(c)PP-DocLayoutV3} 
    \end{minipage}
    \hfill
    \begin{minipage}[b]{0.44\textwidth}
        \centering
        \includegraphics[width=\textwidth]{figures/layout_readingorder5.pdf}
        \centerline{(d)Ours} 
    \end{minipage}

    \caption{\textbf{Visualization of insurance document case.}}
    \label{fig:appendix_four_figures}
\end{figure}

\section{GRPO Alignment Improvements}
\label{app:grpo_alignment}
\FloatBarrier

\begin{figure*}[hbpt]
    \centering
    \includegraphics[width=\textwidth]{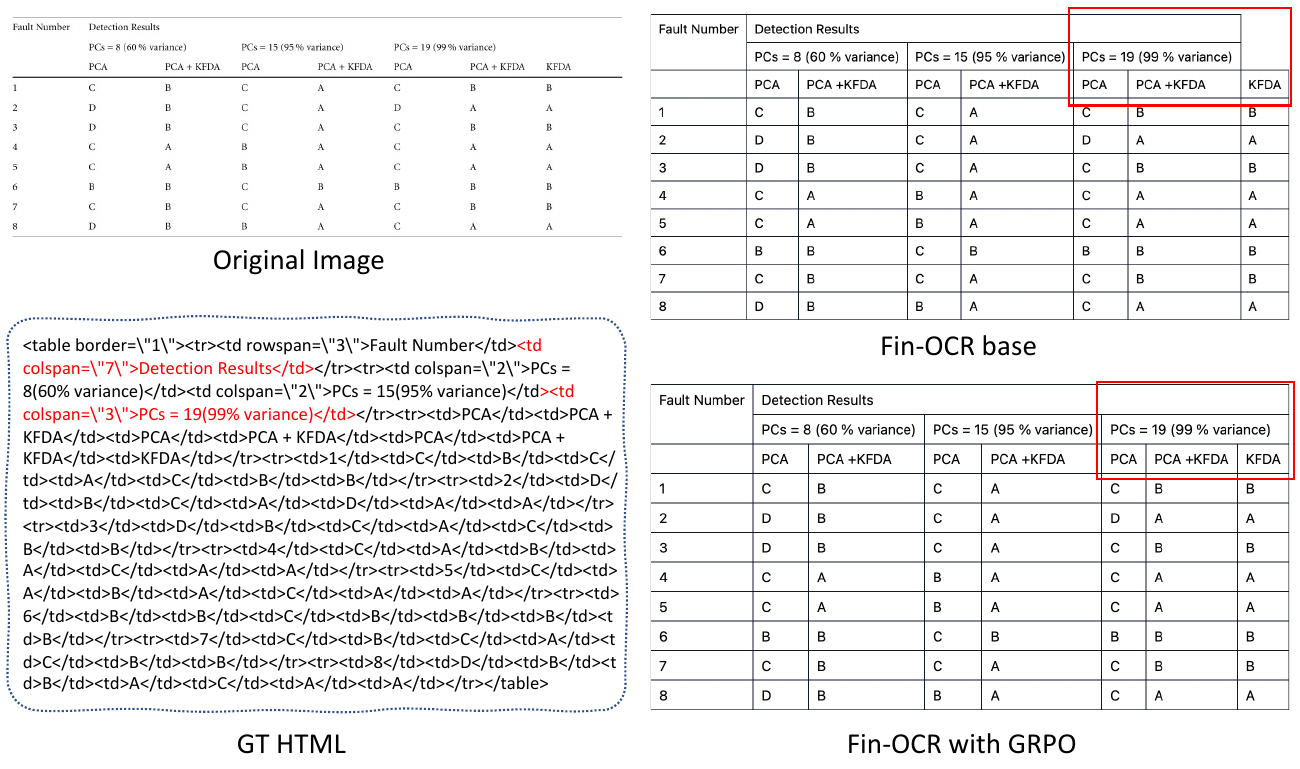}
    \caption{\textbf{Qualitative comparison of table parsing results before and after GRPO.} GRPO improves row/column alignment on complex tables, with particularly noticeable gains in the last few rows and columns.}
    \label{fig:grpo_alignment_example}
\end{figure*}

\appendix

\end{document}